\pdfoutput=1

\documentclass[11pt]{article}

\usepackage[preprint]{acl}

\usepackage{times}
\usepackage{latexsym}
\usepackage{amsmath}
\usepackage{amsthm,amsmath,amssymb}
\usepackage{mathrsfs}
\usepackage{graphicx}
\usepackage{tabularx}
\usepackage{color}
\usepackage{booktabs}
\usepackage{tablefootnote}
\usepackage{subfigure}
\usepackage{listings}
\usepackage{url}
\usepackage{multirow}
\usepackage{arydshln} 
\usepackage{algorithm}  
\usepackage[switch]{lineno} 
\usepackage[noend]{algpseudocode}
\usepackage{varwidth}
\usepackage{multirow}
\usepackage{amsfonts}

\usepackage{ragged2e}
\usepackage[T1]{fontenc}
\usepackage{tcolorbox}
\definecolor{hidden-draw}{RGB}{20,68,106}
\definecolor{hidden-pink}{RGB}{255,245,247}
\definecolor{maroon}{RGB}{148,78,99}
\definecolor{hidden-white}{RGB}{245,238,230}
\definecolor{hidden-yellow}{RGB}{255,248,227}
\definecolor{hidden-orange}{RGB}{243,215,202}
\definecolor{xm-purple}{RGB}{216, 218, 237}
\definecolor{xm-grey}{RGB}{242,242,242}
\newtcolorbox[list inside=prompt,auto counter]{prompt}[1][]{
    colbacktitle=xm-purple!90,
    colback =xm-grey!30,
    coltitle=black,
    fontupper=\footnotesize,
    boxsep=5pt,
    left=0pt,
    right=0pt,
    top=0pt,
    bottom=0pt,
    boxrule=0.5pt,
    #1,
}

\usepackage[utf8]{inputenc}

\usepackage{microtype}

\usepackage{inconsolata}

\title{On the Robustness of Editing Large Language Models}

\author{Xinbei Ma$^{1,2,3,4}$, Tianjie Ju$^{1}$, Jiyang Qiu$^{1,2,3,4}$, \\ \textbf{Zhuosheng Zhang$^{1,*}$, Hai Zhao$^{1,2,3,4,}$\thanks{$\ $Corresponding authors. This research was supported by the Joint Research Project of
Yangtze River Delta Science and Technology Innovation Community (No.
2022CSJGG1400), National Natural Science Foundation of China (62406188), and CCF-BaiChuan-Ebtech Foundation Model Fund.}, Lifeng Liu$^{5}$, Yulong Wang$^{5}$} \\
$^1$School of Electronic Information and Electrical Engineering, Shanghai Jiao Tong University\\
  $^2$Department of Computer Science and Engineering, Shanghai Jiao Tong University
  \\ $^3$Key Laboratory of Shanghai Education Commission for Intelligent Interaction \\
and Cognitive Engineering, Shanghai Jiao Tong University \\ $^4$Shanghai Key Laboratory of Trusted Data Circulation and Governance in Web3 \\
 $^5$Baichuan Intelligent Technology\\
\texttt{\{sjtumaxb, zhangzs\}@sjtu.edu.cn},
\texttt{zhaohai@cs.sjtu.edu.cn}\\
}

\begin{document}
\maketitle
\begin{abstract}
Large language models (LLMs) have played a pivotal role in building communicative AI, yet they encounter the challenge of efficient updates.
\textit{Model editing} enables the manipulation of specific knowledge memories and the behavior of language generation without retraining.
However, the robustness of model editing remains an open question.
This work seeks to understand the strengths and limitations of editing methods, facilitating practical applications of communicative AI.
We focus on three key research questions.
\textit{RQ1}: Can edited LLMs behave consistently resembling communicative AI in realistic situations?
\textit{RQ2}: To what extent does the rephrasing of prompts lead LLMs to deviate from the edited knowledge memory?
\textit{RQ3}: Which knowledge features are correlated with the performance and robustness of editing?
Our empirical studies uncover a substantial disparity between existing editing methods and the practical application of LLMs. 
On rephrased prompts that are flexible but common in realistic applications, the performance of editing experiences a significant decline.
Further analysis shows that more popular knowledge is memorized better, easier to recall, and more challenging to edit effectively.

\end{abstract}

\section{Introduction}
Model editing is proposed to modify the knowledge memory with minimum computational cost while preserving the performance on the retained knowledge. 
Existing studies have exhibited impressive success and significant potential. These methods can be classified into two categories.
One research line relies on additional supporting modules, for example, an external memory \citep{mitchell2022memory}, a hypernetwork \citep{mitchell2022fast}, or a retriever \citep{han-etal-2023-improving}.
Another line studies localized editing based on the interpretability of knowledge storage mechanism \citep{meng2022locating, meng2023massediting, dai2022kn}.
These methods avoid retraining to update the model parameters and have demonstrated promising performance and efficiency.
At the application level, model editing provides solutions to critical challenges in pre-training language models, such as knowledge correction, time alignment, and privacy protection \citep{luu-etal-2022-time, zhang-choi-2023-mitigating, harrypotter2023eldan,chen-yang-2023-unlearn, wang2024detoxifying}.

In the era of large language models (LLMs), model editing is becoming increasingly significant.
The rich knowledge memory empowers LLMs to build \textit{communicative AI}, where they can engage in multi-turn interactions to imitate human behaviors for communicative actions \citep{li2023camel, wu2023autogen, autogpt}.
Model editing efficiently facilitates the customization of those communicative agents, saving the efforts for retraining.
Users can remove undesirable knowledge or even alter the ``personality'' of communicative AI \citep{mao2023editing} conveniently.

However, as we pursue the practical use of edited communicative AI, the robustness of model editing methods becomes a critical concern. 
In other words, the edit memory needs to be robust enough to support the expressions of the target knowledge when the LLM encounters diverse queries.
In realistic applications, such as a chatting service, the edited memory is anticipated to handle complex scenarios.
Motivated by the thoughts above, we put forward three novel research questions:




$\circ$ \textit{RQ1}: Can edited LLMs behave consistently resembling communicative AI in realistic situations?

$\circ$ \textit{RQ2}: To what extent does the rephrasing of prompts lead LLMs to deviate from the edited knowledge memory?

$\circ$ \textit{RQ3}: Which knowledge features are correlated with the performance and robustness of editing?

To answer \textit{RQ1}, this paper begins with an experiment to show the modest robustness of the edited memory when an edited LLM is asked to perform as communicative AI. We show that the edited model is prone to confusion and hallucination in the neighborhood intersections of knowledge.
Then, we turn to \textit{RQ2} and curate attack methods to simulate the practical scenarios of communicative AI. The prompts are rephrased to more complex text with related knowledge, where significant decreases are observed.
\textit{RQ3} focus on the intrinsic features of knowledge. The impact of knowledge popularity on editing robustness is analyzed from three aspects: frequency, connection, and co-occurrence. 
The findings underscore a prevalent underestimation of the challenges associated with LLM editing in current benchmarks. Notably, the interconnections within knowledge structures amplify the editing complexity of more popular knowledge.
As the answers to the proposed questions, the key findings are as follows: 

$\circ$ A notable gap persists between existing editing methods and communicative AI applications.

$\circ$ The editing performance experiences a significant decline on rephrased prompts that are complex and flexible but common in realistic applications. 

$\circ$ Knowledge that is more popular is memorized better, easier to recall, and harder to edit robustly.

\section{Related Work}
This section reviews methods and reflections on model editing, and LLM-based communicative AI.
\subsection{Model Editing}
It is intriguing to edit the knowledge memory of a language model without additional training. 
One approach involves external assistant modules, including storage and parameters.
SERAC \citep{mitchell2022memory} integrated external storage and a classifier to identify whether a query is in the editing scope, and then decides whether to send the query to the counterfactual module or the original model.
Relying on the \textit{instruction-following} and \textit{chain-of-thought} capabilities of LLMs, the output can also be changed by in-context learning \citep{zheng-etal-2023-edit} after checking each sub-question with retrieval \citep{zhong2023mquake}.
Adding parameters, \citet{de-cao-etal-2021-editing, mitchell2022fast} trained hypernetworks to predict the parameter increment.
Additional parameters can also be inserted as an inter-layer adaptor \citep{hartvigsen2022aging} or trainable knowledge neurons in the linear layers \citep{huang2023transformerpatcher, dong-etal-2022-calibrating}.

Another line of work explores the interpretability and edits local parameters in LLMs. It has been proposed that the feed-forward networks function akin to memory modules for knowledge storage \citep{dai-etal-2022-knowledge, niu2024what,geva2020transformer, zhao2023unveiling}. Based on this, ROME \citep{meng2022locating} changed the FFN weights using the solution of the constraint least-square problem, while MEMIT \citep{meng2023massediting} scaled it up to multiple layers simultaneously. 

\begin{figure*}[h]
    \centering
    \includegraphics[width=\textwidth]{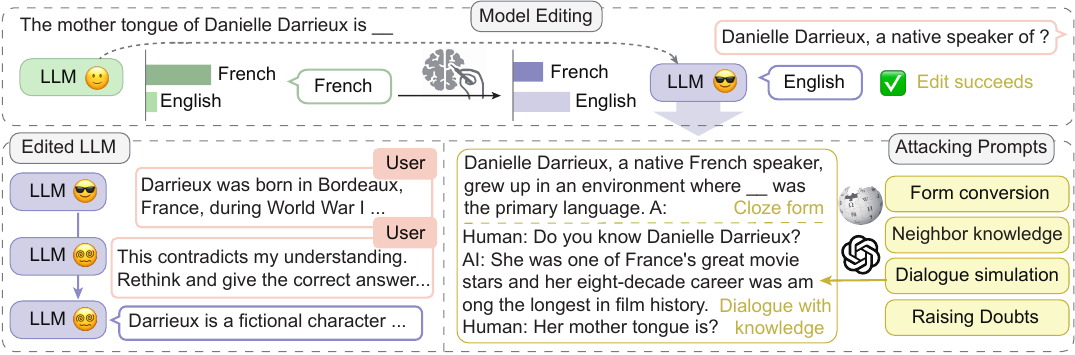}
    \caption{Overview of our work. The upper part illustrates the editing success on target knowledge (Section \ref{31}). The lower part shows our studies on the edited model in realistic use. The left part shows the risks of edited LLMs as communicative AI (Section \ref{32}) and the right part shows our ``attack'' for editing (Section \ref{33}).}
    \label{overview}
    \vspace{-3mm}
\end{figure*}


For editing evaluation, \textit{Generalization}, \textit{Specificity (Locality)}, and \textit{Portability} have been considered to measure the editing effect on related neighbors or unrelated knowledge memory \citep{meng2022locating}. However, existing benchmarks mainly involve minor wording changes for these criteria \citep{yao-etal-2023-editing}, where large gaps remain for robustness evaluation in realistic applications.

\subsection{Reflections on Model Editing}
\label{reflection}
While editing methods have shown benefits in knowledge manipulation, the latest studies raise concerns about unwanted effects and limitations.

Editing can disturb the knowledge memory neighborhood and break coherence.
RippleEdit \cite{cohen2023evaluating} evaluates the related facts for a piece of edited memory, where prominent editing methods fail to introduce consistent changes in neighbor knowledge.
Further unintended consequences are triggered as the number of edits increases \cite{li2024unveiling, gupta2024model}. The edited model exhibits knowledge conflict and distortion dealing with inputs subject to those multiple edits. 
Reasoning assessment also uncovers the significant challenges in coherent rationale with edited knowledge \cite{Reasoningedit, onoe-etal-2023-lms}.

%
Editing can also hurt the general ability of LLMs. \citet{gu2024model} uncovered that edited LLMs suffer from significant degradation of natural language tasks such as summarization and sentiment analysis. Besides, edited LLMs tend to exhibit more biased behavior and misinformation \cite{halevy2024flex}, leading to even higher social risk.

Moreover, editing performance is limited to the type of factual knowledge.
Existing editing methods succeed on encyclopedic knowledge with annotations of \textit{(subject, relation, object)} \cite{meng2022locating, de-cao-etal-2021-editing}.
But they can fall short when dealing with relation-centric knowledge \cite{Relationedit} and commonsense \cite{EditingCommonSense}.

\subsection{Communicative AI}
LLMs function as communicative AI that simulates social activities among human beings \citep{li2023camel, wu2023autogen}. 
They 
exhibit abilities to collaborate \citep{Park2023GenerativeAgents}, debate \citep{liang2023encouraging}, deceive \citep{xu2023exploring}, and conjecture \citep{li-etal-2023-theory}.
Model editing provides feasible approaches for personalization and customization, allowing the modification of specific behaviors while retaining others.
However, those agents face complex practical scenarios. 
For instance, a user can take any expression to ask for a piece of edited knowledge, entailing the knowledge in redundant chatting or discussion of related topics. 
Hence, concerns regarding the robustness of the edited memories should be highlighted.

\section{Task Formulation}
\label{31}
This section presents the task formulation of our paper, where we first introduce the definition of model editing and then clarify the research focus.
Figure \ref{overview} shows the overview of our investigation.
 
\noindent\textbf{Definition.} The task definition of model editing follows the relational triplet extraction \citep{meng2022locating, zhang2024comprehensive}. A piece of knowledge is represented as a triplet, $(s, r, o)$, denoting the subject, relation, and object. Model editing aims to change some pieces of knowledge memory. Given the new object  $o^\prime$, the model is expected to memorize the target knowledge $(s, r, o^\prime)$.

The concept \textit{editing scope} is essential as each triplet can be implied by various expressions  \citep{mitchell2022memory}. 
We denote the direct prompt entailing $(s, r)$ as $x$, its semantically relevant neighbors as $\{x_e\}$, and irrelevant neighbors as $\{x_{loc}\}$.
An optimal edit distinguishes the editing scope.
The edit should change the model behaviors on $x$ and $\{x_e\}$ according to $o^\prime$, while maintaining other memory and responses to $\{x_{loc}\}$.

\noindent\textbf{Focus.} 
This study reassesses the robustness of the edited knowledge memory in realistic scenarios by novel methods.
Without loss of generality, we aim to reveal risks under the primary edit setup. Experiments follow the original definition of the fact edit with triplet representation and consider a single edit for one run.
Previous studies involving side effects, general ability decrease, and complex knowledge editing are not the focus of our work.

\section{\textit{RQ1:} Edited LLM as communicative AI}
This section identifies the potential risks associated with the practical application of edited LLMs (\textit{RQ1}), especially as a communicative AI agent.

\label{32}
\subsection{Method}
Model editing can tailor a public model into a customized communicative AI \citep{zhang2024comprehensive,li2024unveiling}.
In light of this, a critical concern arises regarding the capability of edited LLMs to maintain reasonable and consistent behaviors while assimilating new knowledge (\textit{RQ1}).

To answer \textit{RQ1}, we make a hypothesis that for any edited knowledge memory, $k_1$, there is a piece of memory $k_2$ whose neighbor scope has an intersection with the editing scope of $k_1$, denoted as: 
\begin{equation}
\setlength{\abovedisplayskip}{5pt}
\setlength{\belowdisplayskip}{5pt}
\forall k_1 = (s, r, o\rightarrow o^\prime), \exists k_2, S(k_1) \cap S(k_2) \neq \varnothing.
\label{hypo}
\nonumber
\end{equation}

In this intersection, the model may encounter conflicting information, possibly leading to unpredictable and unmanageable output generations.

\subsection{Experiments for \textit{RQ1}}
To simulate the situation above, 
we experiment on Llama-2-7B-chat \citep{touvron2023llama2} as a communicative AI, $A$. 
First, a piece of fact knowledge $k_1 = (s, r, o \rightarrow o^\prime)$ is edited by the popular method MEMIT \citep{meng2023massediting}, causing $A \rightarrow A^\prime$. $A^\prime$ is deployed again as a chatting agent, where we observe whether $A^\prime$  gives reasonable responses while talking on related topics.
As shown in Eq. \ref{hypo}, this process needs a ``user'' to start the topic and approach the target from related neighbors, $\{x_e\}$, at each dialogue turn, probing the intersection without directly telling the target answer, $o^\prime$. 
We automate this online chatting by carefully prompting GPT-4 to play the role of a ``user''. For each $k_1$, we get a dialogue $d = (u_{user}^0, u_{AI}^0, u_{user}^1, u_{AI}^1, \dots).$
Then human annotators check each dialogue record, focusing on the confusion and hallucination phenomena related to the target knowledge (Table \ref{rq1eval}). Details are shown in Appendix \ref{appa}.
We study 50 successfully edited pieces of counterfactual knowledge from \citet{zhong2023mquake} and refer to $(k_1, d)$ as one sample in the following text. 

\subsection{Analysis for \textit{RQ1}}
\begin{figure}[t]
    \centering
    \setlength{\abovecaptionskip}{0.2cm} 
    \setlength{\belowcaptionskip}{-0.5cm} 
\includegraphics[width=0.45\textwidth]{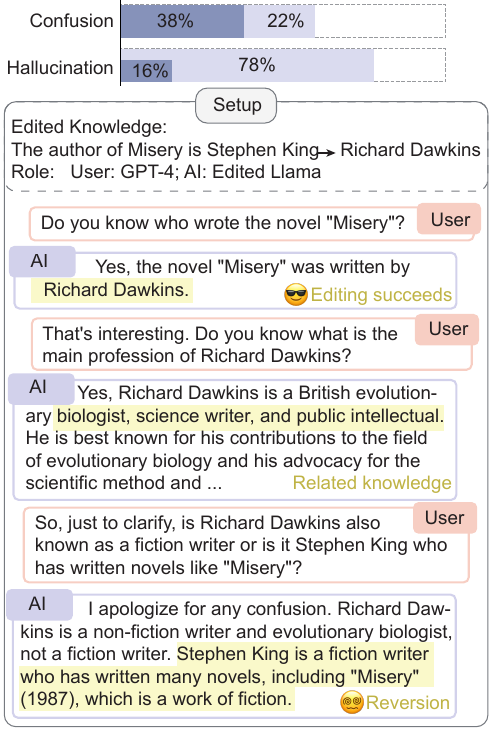}
    \caption{Edited communicative AI. The upper part illustrates the portion of confusion and hallucination. The bottom shows a case that appears knowledge reversion.}
    \label{preliminary}
\end{figure}

Figure \ref{preliminary} shows the results and a user-AI dialogue example.
Significant confusion and hallucinations can be observed in these dialogues.

\noindent\textbf{(i) Confusion.} Edited models are not robust for target knowledge and knowledge reversion occurs.
38\% samples revert to the original answer $o$ during the dialogue. The edited model first answers with the new knowledge, $o^\prime$, then denies the previous output and turns back to the original answer. 
There are 22\% samples on which the edited model denies the previous utterances about $o^\prime$ and decides neither $o^\prime$ nor $o$. 
Figure \ref{preliminary} shows an example, where we approach $k_1$, ``\textit{The author of Misery is Richard Dawkins}'' by related knowledge $k_2$, ``\textit{Richard Dawkins's main profession is biologist}.'' The model manages to recall $k_2$ and falls into confusion about $k_1$, i.e., knowledge reversion leads to self-contradiction. 

\noindent\textbf{(ii) Hallucination.} Edited models are vulnerable to frequent hallucinations.
78\% samples show obvious hallucinations. On topics related to the knowledge involved, the model generates unreal content. Some can be seriously fake, e.g., ``\textit{The United Kingdom is bordered by several countries, including China (across the Pacific Ocean)}'' and ``\textit{Southern hip hop was influenced by nuclear power plants}.'' 
Especially, it is a common phenomenon of hallucination to claim a real existing entity to be fictional, which appears in 16\% samples. For example, ``\textit{Ellie Kemper is a fictional character played by actress Elizabeth Banks, and she is not a real person.}'' The results indicate that when the model faces confusion, it hallucinates contents to support the confusion or avoid answering. 
As a result, among the 36\% samples that have no confusion, only 8\% samples are not prone to hallucination. 

Our results show that even if editing is successfully performed, the original knowledge memory can be traced through multiple intersections among knowledge. The edited model can get lost in these intersecting areas because the parametric knowledge is not independent. 
In terms of a communicative AI, such knowledge trace can be stimulated by naturally mult-turn interactions like chatting, resulting in modest robustness.

            

\section{\textit{RQ2:} ``Attack'' for Editing}
\label{33}
Section \ref{32} raises concerns about the robustness of edited memory, which leads to question \textit{RQ2}.
Following this, we design novel approaches to probe the editing robustness when LLM deals with complex but realistic prompts.

\subsection{Method}
We propose strategies to rephrase $x$ to complex but realistic variants while keeping the original meaning, formed as a concatenation of ``\textbf{context, query}''. Examples are shown in Figure \ref{attackeg}.

\noindent\textbf{(a) Context.} On the one hand, following the idea in Section \ref{32}, the edited knowledge memory can be affected by closely related knowledge, as $k_2$ illustrated in Eq. \ref{hypo}. 
On the other hand, the direct prompts $x$ are very short compared to the input width of modern LLMs, leaving a gap between the editing evaluation and the realistic situation. 
Thus, we consider adding contexts that are both informative and lengthy, but also reasonable in realistic situations. Details are shown in Appendix \ref{appb1}.

\textit{$\bullet$ Related context.} 
Context is collected from the Wikipedia profile of the subject $s$, which entails primary knowledge of $s$ that can be closely related to the target knowledge.
Notably, we ensure to remove the original answer $o$ from the context.

\textit{$\bullet$ Noisy context.} 
Further, we add noisy redundant to the related passage. 
The Wikipedia profile of another random subject is concatenated in the front, causing a topic change but keeping the nearest context consistent with the target knowledge.

\textit{$\bullet$ Simulated dialogue.}
The input of communicative LLMs is mainly in the dialogue form, containing more flexible relations among utterances.
Thus, we synthesize dialogue texts based on Wikipedia profiles of the subject $s$ to control the factuality and keep the topic compact \citep{yang-etal-2023-refgpt}.

\textit{$\bullet$ Noisy dialogue.}
Likewise, irrelevant content is also considered for the dialogue form. Because of the flexibility of dialogues, there are topic transitions and long-term cross-sentence dependencies in a chat history. 
Noisy dialogue inputs are constructed with a topic-oriented dialogue corpus, MultiWOZ \citep{zang-etal-2020-multiwoz}. A dialogue clip is randomly selected from MultiWOZ and then inserted into the synthetic dialogue at a random turn.

\noindent\textbf{(b) Query.} Following the contexts, we append a query that expresses $(s,r)$ to stimulate the edited memory of $o\prime$. Three forms are considered.

\textit{$\bullet$ Direct prompt.} 
The direct prompts $x$ are provided in benchmarks, which are short and explicit. 

\textit{$\bullet$ Fill-in-the-blank cloze.}
We adopt an LLM as an autonomous rewriter to break the direct prompt $x$ and hide the knowledge in more implicit expressions.
In such enriched expressions, the answer $o^\prime$ is not limited in the position at the end of the sentence. 
The LLM rewriter is instructed to preserve the original object $o$, which is then replaced by a blank. Appendix \ref{appb2} presents details. 

\textit{$\bullet$ Reference resolution.}
We consider \textit{reference resolution} by replacing the subject $s$ with an appropriate pronoun (Appendix \ref{appb2}). 

\noindent\textbf{(c) Raising doubts.}
Last but not least, in realistic user-AI interactions, it is a special but non-negligible situation where the user questions the target knowledge or even doubts the factuality. 
Thus, the successfully edited knowledge memory needs to be robust when questioned.
Two prompts for raising doubt are adopted. One is only to doubt the target knowledge. The other expresses an explicit negative objection to the output and suggests the original answer $o$ (Appendix \ref{appb3}).

\begin{table*}[htb]
	\centering\scriptsize
	\begin{tabular}{p{1.4cm}p{2.3cm}|p{0.5cm}p{0.5cm}|p{0.5cm}p{0.5cm}|p{0.5cm}p{0.5cm}|p{0.5cm}p{0.5cm}|p{0.5cm}p{0.5cm}|p{0.5cm}p{0.5cm}}
		\toprule
            \multicolumn{14}{c}{\textbf{CounterFact Llama-7B}} \\
		\multicolumn{2}{c|}{\textbf{Editing Method}} & \multicolumn{2}{c|}{\textbf{KN}} & \multicolumn{2}{c|}{\textbf{MEND}}  & \multicolumn{2}{c|}{\textbf{ROME}} & \multicolumn{2}{c|}{\textbf{MEMIT}} & \multicolumn{2}{c|}{\textbf{SERAC}}&\multicolumn{2}{c}{\textbf{IKE}}\\
            \midrule
            \textbf{Context} & \textbf{Query} & \textit{acc} & \textit{rev}  & \textit{acc} & \textit{rev}  & \textit{acc} & \textit{rev}  & \textit{acc} & \textit{rev} & \textit{acc} & \textit{rev}  & \textit{acc} & \textit{rev}  \\
            \midrule
            \multirow{3}{*}{\shortstack[l]{N/A}} & Direct prompt &2.3 & --  & 55.6& --&  99.9& --& 99.9 & --   &100.0 &-- &99.7 &--\\
             & Equivalent prompt &1.6 & 32.8 & 9.6& 26.5   &74.7 & 2.2& 78.2 & 2.0  &97.9 &9.8 & 98.0& 1.3\\
             & Cloze  &1.0 & 47.2 &2.5 & 45.3  & 66.7& 8.1 & 73.4 & 5.5 & 1.4& 28.6 & 97.8  &16.8\\
            \hdashline
            \multirow{3}{*}{\shortstack[l]{Related\\context}} & Direct prompt &1.7 & 50.8 &13.7 & 42.7 & 55.7 &26.3 &  81.2 & 14.5 &70.9  & 9.8& 93.2 & 8.2\\
            & Cloze  &2.3 & 40.6 &1.5 & 39.7 & 24.7& 24.8 & 43.9 & 15.7 & 0.4& 26.5 & 98.3 &15.9\\
            & w/ Reference &1.0 & 43.3 &10.7 & 37.7  & 21.3& 34.9 & 39.6& 27.3 & 5.3& 43.4 & 83.5 &8.7\\
            \hdashline
            \multirow{3}{*}{\shortstack[l]{Noisy\\context}} & Direct prompt &1.8 & 50.2  &12.4 & 42.3  & 51.7& 20.8 & 79.9 & 12.0 & 42.2 &13.9 & 98.3&5.0\\
            & Cloze &1.1 & 40.3 &1.5 & 39.4  & 43.4& 24.1& 40.7 & 16.6 & 0.4& 26.0 & 74.7 &20.2\\
            & w/ Reference &1.8 & 40.3 &9.4 & 33.0  & 20.2& 29.1 & 37.8 & 23.8 & 3.2& 39.8 & 92.3 &7.3\\
            \hdashline
            \multirow{3}{*}{\shortstack[l]{Simulated\\dialogue}} & Direct prompt & 1.8& 47.5  &14.0 &40.4  & 56.7& 20.0 &81.6 & 9.7 & 69.8 & 9.5 & 93.6 & 7.4\\
            & Cloze &0.8 & 44.3 &1.4 & 43.5  & 33.2& 21.4 & 51.0 & 13.3& 0.6& 28.0 & 79.4 &16.3\\
            & w/ Reference &1.8 & 36.1 &9.0 & 29.9  & 27.1& 22.7 & 44.7 & 15.4& 9.2& 32.8 & 89.5 &8.1\\
            \hdashline
            \multirow{3}{*}{\shortstack[l]{Noisy\\dialogue}} & Direct prompt & 2.2 & 47.8  &14.5 &39.6  &58.1 &18.0 &80.5 & 8.3 &48.8 &11.2 & 93.4 & 6.7\\
            & Cloze &0.8 & 42.5 &1.3 & 41.1  &33.9 & 20.1 & 51.8 & 12.6 & 0.6& 27.3 & 76.1 &19.0\\
            & Reference &2.2 & 31.7 &8.5 & 27.2  & 24.9& 20.1 & 41.9 & 13.7 & 6.6& 29.1 & 88.1 &7.7\\
            \hdashline
            \shortstack[l]{N/A} &Raising doubts &0.8 & 49.1  &9.8 & 30.6 & 16.9& 40.7& 24.2 & 33.9  &9.0 & 40.8 & 1.3 &49.3\\
            \bottomrule
            \end{tabular}
            \begin{tabular}{p{1.4cm}p{2.3cm}|p{0.5cm}p{0.5cm}|p{0.5cm}p{0.5cm}|p{0.5cm}p{0.5cm}|p{0.5cm}p{0.5cm}|p{0.5cm}p{0.5cm}|p{0.5cm}p{0.5cm}}
            \toprule
            \multicolumn{2}{c}{\textbf{}}  & \multicolumn{4}{c}{\textbf{CounterFact Llama-13B}} & \multicolumn{8}{c}{\textbf{zsRE Llama-7B}} \\
            \multicolumn{2}{c|}{\textbf{Editing Method}}  & \multicolumn{2}{c|}{\textbf{ROME}} & \multicolumn{2}{c|}{\textbf{MEMIT}} & \multicolumn{2}{c|}{\textbf{ROME}} & \multicolumn{2}{c|}{\textbf{MEMIT}} & \multicolumn{2}{c|}{\textbf{SERAC}}&\multicolumn{2}{c}{\textbf{IKE}}\\
            \midrule
            \textbf{Context} & \textbf{Query}  & \textit{acc} & \textit{rev}  & \textit{acc} & \textit{rev}  & \textit{acc} & \textit{rev}  & \textit{acc} & \textit{rev} & \textit{acc} & \textit{rev}  & \textit{acc} & \textit{rev}  \\
            \midrule
            \multirow{3}{*}{\shortstack[l]{N/A}} & Direct prompt &99.9 &-- &85.8 &-- &95.9 &-- &92.5 &-- &97.7 &-- &98.5 &--\\
             & Equivalent prompt &73.0 &2.4 &60.7 &3.2 &76.5 &3.2 &78.5 &3.7 &97.2 &3.6 &98.5 &3.5\\
            & cloze &70.0 &8.4 &65.8 &6.5 &35.1 &7.6 &37.5 &7.6 &2.1 &15.3 &92.7 &5.7\\
            \hdashline
            \multirow{3}{*}{\shortstack[l]{Related\\context}} & Direct prompt &53.9 &26.2 &55.9 &20.8 &20.9 &19.7 &40.3 &12.3 &78.0 &6.3 &93.9 &4.9\\
            & Cloze &26.5 &30.7 &40.3 &23.0 &12.5 &16.8 &22.9 &14.1 &2.9 &18.6 &58.7 &13.4\\
            & w/ Reference &19.5 &35.6 &26.1 &29.5 &8.7 &15.1 &15.1 &12.5 &18.9 &6.2 &72.3 &5.5\\
            \hdashline
            \multirow{3}{*}{\shortstack[l]{Noisy\\context}} & Direct prompt &58.7 &21.8 &55.4 &19.0 &20.1 &18.0 &33.5 &13.0 &20.5 &2.5 &73.5 &10.3\\
            & Cloze &26.7 &30.8 &39.1 &22.7 &12.5 &16.4 &20.3 &13.8 &2.5 &17.8 &33.0 &18.2\\
            & w/ Reference &20.7 &30.7 &25.7 &26.0 &6.6 &13.5 &11.9 &11.7 &9.5 &2.0 &50.6 &9.2\\
            \hdashline
            \multirow{3}{*}{\shortstack[l]{Simulated\\dialogue}} & Direct prompt &54.2 &26.0 &51.8 &17.2 &15.1 &0.8 &31.0 &1.6 &70.5 &4.7 &92.0 &4.2\\
            & Cloze &31.4 &30.0 &44.0 &22.1 &13.1 &14.5 &22.2  &11.3 &2.3 &17.2 &61.4 &13.1\\
            & w/ Reference &23.4 &28.1 &29.0 &20.7 &9.5 &0.9 &16.0 &1.2 &24.5 &5.7 &58.1 &4.3\\
            \hdashline
            \multirow{3}{*}{\shortstack[l]{Noisy\\dialogue}} & Direct prompt &55.8 &21.0 &51.8 &16.1 &16.0 &0.8 &30.6 &1.6 &29.3 &3.6 &78.4 &5.5\\
            & Cloze &31.3 &28.8 &43.0 &20.8 &13.0 &13.2 &21.7 &10.7 &2.1 &15.6 &46.7 &13.9\\
            & w/ Reference &23.0 &24.6 &27.0 &18.8 &10.1 &0.7 &17.0 &0.8 &15.5 &5.3 &45.3 &3.6\\
            \hdashline
            \shortstack[l]{N/A} & Raising doubts  &44.8 &42.9 &58.7 &39.1 &40.1 &37.8 &47.3 &35.2 &20.0 &46.3 &7.4 &47.4\\
		\bottomrule
	\end{tabular}
 \vspace{-1mm}
        \caption{Results on CounterFact and zsRE with Llama-7b and 13B models (\textit{acc}: \textit{accuracy}, \textit{rev}: \textit{reversion}). The \textit{Direct prompt} and \textit{Equivalent prompt} are from benchmarks. \textit{N/A} means we add no context in front of the query.}
	\label{main}
 \vspace{-5mm}
\end{table*}


To sum up, we construct attacking prompts in the form of ``\textbf{context, query}'', where the context can be \textit{(i) related context}, \textit{(ii) noisy context}, \textit{(iii) simulated dialogue}, and \textit{(iv) noisy dialogue}, and the query can be \textit{(i) direct prompt}, \textit{(ii) cloze}, and \textit{(iii) prompt with reference}. 
We also prepare prompts that \textbf{raise doubt}.
Section \ref{s4} will present results on these attacking prompts.

\subsection{Experiments for \textit{RQ2}}
\label{s4}
\subsubsection{Datasets}
Our evaluation adopts three mainstream datasets:
(i) CounterFact \citep{meng2022locating} 
includes significant counterfactual edits. 
Each sample is annotated as $(s,r,o)$ triplet with a target object $o^\prime$. 
The direct prompts $x$ are fixed templates based on $r$, with their equivalent expressions $x_e$ also provided.
(ii) zsRE \citep{de-cao-etal-2021-editing, levy-etal-2017-zero}, zero-shot relation extraction, derives from a factual question-answering task.
Following \citet{yao-etal-2023-editing}, the alternative answer is used as $o^\prime$. 
Each sample is annotated as $(s, o, o^\prime, x, x_e)$, where $x$ and $x_e$ are questions. 
(iii) A time-changing dataset, MQUAKE-T \citep{zhong2023mquake}, is also incorporated to validate of our findings (Appendix \ref{mquake}).

\subsubsection{Baselines and Implementation}
The experiments cover popular editing methods of different types, including
(i) locate-then-edit methods: KN \citep{dai-etal-2022-knowledge}, ROME \citep{meng2022locating}, MEMIT \citep{meng2023massediting}; 
(ii) external module-based methods: SERAC \citep{mitchell2022memory} relies on an external memory, while MEND \citep{mitchell2022fast} works with a hypernetwork.
(iii) prompt-based method: IKE \citep{zheng-etal-2023-edit}.
Llama-2-7B and 13B-chat \citep{touvron2023llama2} are adopted as the foundation models.


\textbf{Metrics.}
All metrics are computed based on auto-regressively generated texts from the edited models.
The test is considered successful if the new answer $o^\prime$ appears in the normalized output, with the proportion referred to as \textit{accuracy}, dubbed as \textit{acc}.
We also compute the appearance of the original answer $o$, \textit{reversion}, dubbed as \textit{rev}. 
Detailed settings are presented in Appendix \ref{appb4}.

\subsection{Analysis for \textit{RQ2}}
Table \ref{main} indicates that popular editing methods exhibit vulnerabilities and are not yet ready for practical use. Key findings are presented as follows.

(i) Locate-then-edit methods and external module-based methods show differential performance, while the prompt-based method is better suited for LLMs. Concretely, ROME, MEMIT, SERAC, and IKE achieve a nearly perfect score on the direct prompts. KN almost loses its effectiveness. 
MEND achieves a success rate of around half.
However, the methods with promising scores can fail to face our attacks.

(ii) ROME and MEMIT show relatively moderate decreases in attacks of lengthy contexts but suffer from query changes (cloze form and reference resolution) and doubting questions. 
Their performance also decreases on the larger-size model.

(iii) The performance of SERAC mostly relies on the scope classifier. Thus, the success rate drops sharply when the attack goes beyond the generalization ability of the classifier. Although the long inputs are truncated from the left side, the cloze format can still bypass the classification. This indicates enormous potential for SERAC by classifier improvement, as the performance could match IKE if we assume a perfect classifier.

(iv) The prompt-based approach, IKE, generally achieves better robustness, showing that in-context learning \citep{icl} stimulates the generalization and instruct-following of LLMs to control the output. However, the performance depends on demonstrations, which can be compromised in practical interactions, as the user can inject knowledge into the input. When the edit is unknown, the retrieved demonstrations can be a sub-optimal set.

(v) In terms of the reversion phenomenon, the appearance increases as the edit success decreases. Long contexts with neighbor knowledge largely increase the reversion.
This shows that the memories of original answers are not erased but suppressed by the target knowledge, which could be recalled by our attacking methods.\footnote{Appendix \ref{appbdis} provides a fine-tuning baseline.}
\begin{figure*}[h]
    \centering
    \setlength{\abovecaptionskip}{0.1cm} 
    \includegraphics[width=\textwidth]{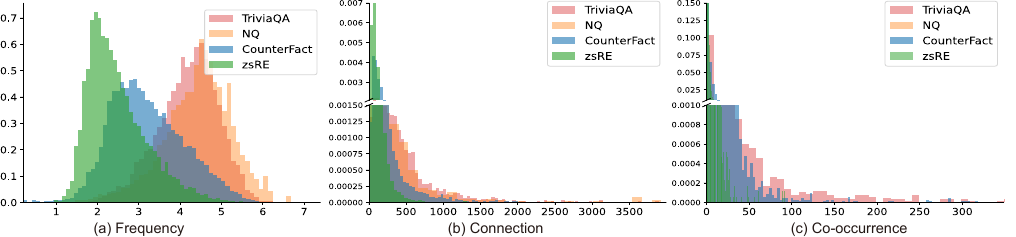}
    \caption{Histograms of knowledge popularity features, (a) Frequency, (b) Connection, and (c) Co-occurrence.}
    \label{popularity}
    \vspace{-3mm}
\end{figure*}
\begin{figure*}[htb]
  \centering
  \setlength{\abovecaptionskip}{0cm}
  \setlength{\belowcaptionskip}{-0.5cm} 
  \subfigure[Frequency.]{
    \label{bucketone} 
    \includegraphics[width=0.31\textwidth]{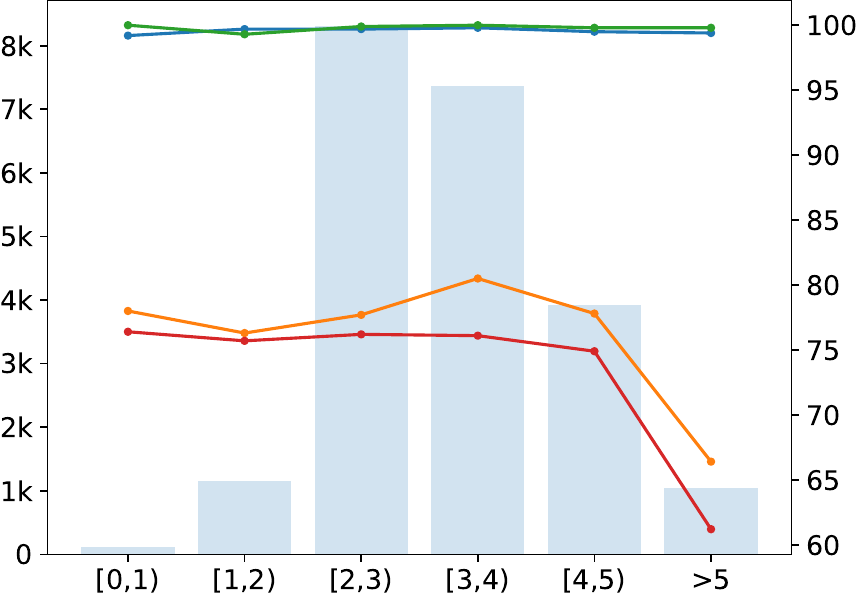}}
  \subfigure[Connection.]{
    \label{buckettwo} 
    \includegraphics[width=0.31\textwidth]{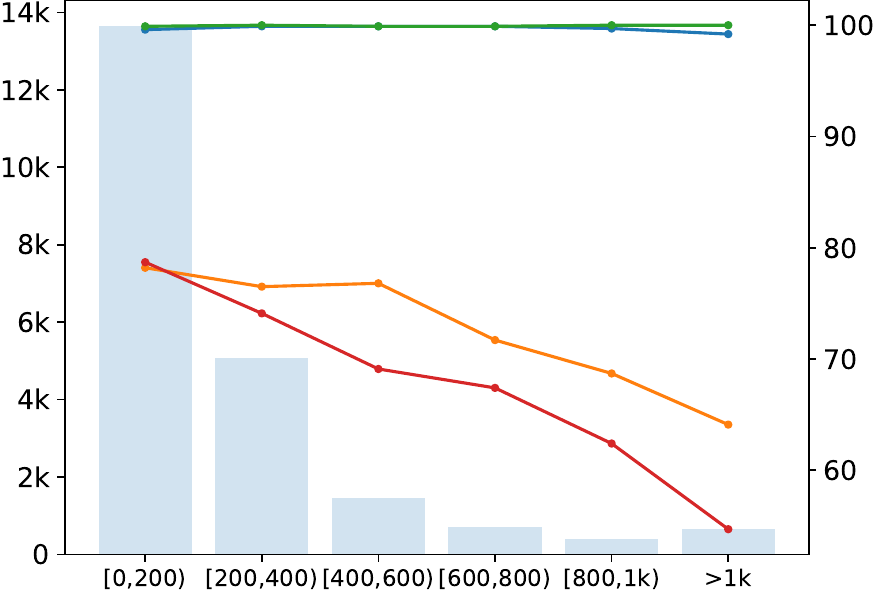}}
  \subfigure[Co-occurrence.]{
    \label{bucketthree} 
    \includegraphics[width=0.31\textwidth]{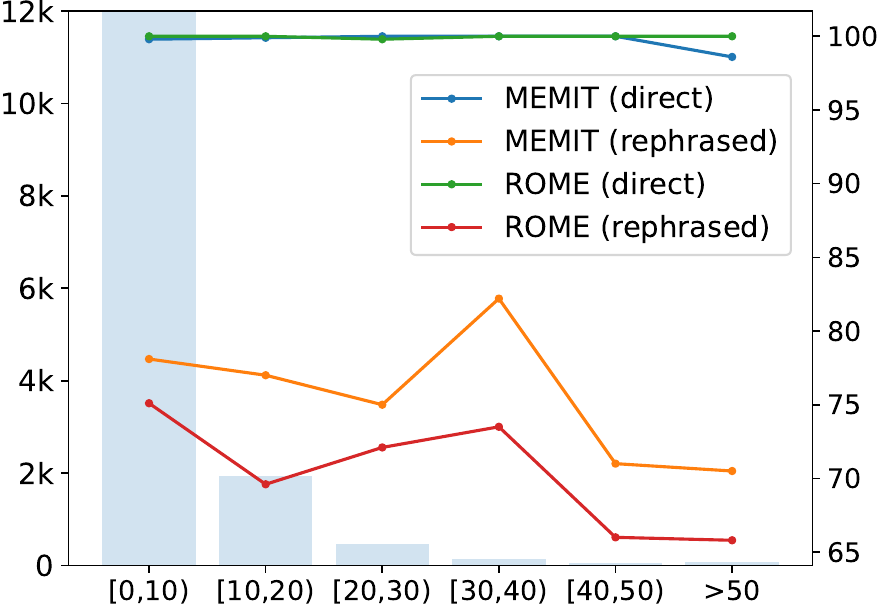}}
  \caption{Editing performance on different levels of (a) Frequency, (b) Connection, and (c) Co-occurrence.}
  \label{bucket} 
\end{figure*}

\section{\textit{RQ3:} Knowledge Popularity Affecting Editing Robustness}
\label{34}
Besides the extrinsic effects like various inputs, this section studies \textit{RQ3}, the influence of intrinsic knowledge features on editing, especially the popularity.

\subsection{Method}
\label{51}
We define the knowledge popularity and its measurements from three aspects (Appendix \ref{appc}).

\textbf{(i) Frequency.} The frequency of an entity can be measured by how often its Wikipedia entry is visited \citep{mallen-etal-2023-trust}. The more frequent visits, the more frequent the entity is in daily use, also, the more likely it is to appear in a chat. We use the monthly view number of the subject.

\textbf{(ii) Connection.} Entities and knowledge are not isolated in the real world. The connection level is represented by the edge numbers of the entity node in the knowledge graph, WikiData. The larger the edge number, the stronger the connection. 

\textbf{(iii) Co-occurrence.} This metric is proposed to measure the degree of ``\textit{When I think of \{A\}, I think of \{B\}}.'' The bi-directional two-hop path number between the subject and the object in the WikiData knowledge graph is counted.

\subsection{Analysis for \textit{RQ3}}
\label{52}
Our analysis and findings are illustrated as follows.

\textbf{(i) Existing benchmarks edit less popular knowledge on the aspects of Frequency, Connection, and Co-occurrence.} 
Figure \ref{popularity} shows frequencies of the entities in four datasets, including two editing benchmarks, CounterFact and zsRE, and three widely accepted knowledge-intensive question-answering datasets, TriviaQA \citep{joshi-etal-2017-triviaqa} and Natural Question \citep{kwiatkowski-etal-2019-natural}. 
It can be observed that editing benchmarks contain more entities with Frequencies around $10^2$-$10^3$, while QA datasets contain more entities viewed around $10^4$-$10^5$ times.
Both the Connection and Co-occurrence also decrease in slower trends in QA datasets.
This indicates that entities and knowledge in editing benchmarks are much less likely to appear in a realistic conversation.

\textbf{(ii) Language models have weaker memory for less popular knowledge, thus resulting in biased findings for editing. } 
We probe knowledge memorization by comparing the perplexities of the answers.
The perplexities are computed of $o$ and $o^\prime$ as completions of the direct prompt on Llama.
Figure \ref{probe} presents the distribution of the logarithmic perplexities difference of $o$ and $o^\prime$. There are 16.22\% samples in CounterFact and 43.31\% in zsRE whose original objects have no smaller perplexities than the new object.

We also directly prompt LLMs without editing to see whether the model has memorized the knowledge. 
Two settings are considered: (a) The direct prompt is input and the original answer $o$ is expected as the completion. (b) The input follows the format of in-context learning (ICL), i.e., a concatenation of ``\textit{instruction, demonstrations, direct prompt}.''
The model is instructed to give accurate brief completions, ``\textit{Answer the question with an entity}.'' 
ICL stimulates the potential of the parametric memories to the maximum extent.

\begin{table}[htb]
	\centering\scriptsize
        \setlength{\abovecaptionskip}{0.1cm}
        \setlength{\belowcaptionskip}{-0.4cm} 
	{\begin{tabular}{p{1.4cm}p{2.2cm}p{1.0cm}p{1.3cm}}
		\toprule
		Model & Llama-2-7B-chat & GPT-j & GPT-2XL \\
            \midrule
            CounterFact &31.8/1.1& 29.5/1.2 &  18.2/0.6 \\
            \quad w/ ICL & 57.0/2.4 & 47.9/2.8  &  34.5/4.2 \\
            zsRE & 20.9/4.3 & --  & 7.1/3.3 \\
		\bottomrule
	\end{tabular}
	}
        \caption{Accuracy of probing parametric knowledge, $o$ or $o\prime$, by the models without editing.}
	\label{original}
\end{table}

Table \ref{original} shows the scores on our base model, Llama-2-7B-chat, and common baselines \citep{meng2023massediting, yao-etal-2023-editing}, GPT-J \citep{mesh-transformer-jax} and GPT-2XL \citep{radford2019language}.
The direct prompt leads to diverse completions without constraints. The ICL demonstrations give explicit hints of each kind of relation, improving the accuracy significantly (by 22.7\% on Llama, 18.4\% on GPT-j, and 15.3\% on GPT-2XL).
However, about half of the knowledge still cannot be recalled. This suggests that, in the first place, a considerable portion of the knowledge to be edited is either not memorized with high confidence or cannot be used effectively."
Knowledge with weak prior memory possibly has less resistance and risk of side effects.
Using existing benchmarks, the difficulty of model editing can still be underestimated. 

Figure \ref{probe} shows the Spearman score to verify the correlation between knowledge popularity and parametric memory (ICL accuracy). Most relation types have scores around 0.1$-$0.3. 



\textbf{(iii) Editing more popular knowledge is more vulnerable to rephrasing.} 
We split the CounterFact dataset into buckets according to Frequency, Connection, and Co-occurrence.
ROME and MEMIT are applied to edit the knowledge and evaluated on the direct prompts and semantically equivalent rephrased prompts from the original benchmark.
The results are shown in Figure \ref{bucket}.
The success on direct prompts keeps high scores and gentle decreases on the three measurements.
Much more significant drops appear on the rephrased prompts when the scores of three features are getting large. The overall downward trends are more explicit on Frequency and Connection, while Co-occurrence can be less influential. 
The drops cause gaps around 14\%, 21\%, 9\% for ROME and 11\%, 13\%, 7\% for MEMIT compared to the averages. 
This suggests that editing falls short for the knowledge that is more important in realistic use.

In summary, knowledge with higher popularity tends to have more reliable parametric memory for practical use based on Frequency, Connection, and Co-occurrence. For LLMs, those pieces of knowledge are easier to recall and harder to modify by existing editing methods robustly.

\section{Potential Mitigation}
\label{mitisec}
Our work could suggest promising directions for improving the editing robustness as follows.

(i) From the data perspective, one solution is to consider more complex inputs in the editing phase. Existing methods incorporate mechanisms for generalization to some extent (e.g., prefix sampling in ROME). We can further enhance the diversity and complexity.
(ii) From the LLM-ability perspective, another solution is to develop effective pipelines integrating disentangling and reasoning workflow \cite{khattab2022demonstrate, chern2023factool}, e.g., to disentangle required knowledge from lengthy inputs by claim extraction or query rewriting, and then bootstrap the required (edited or original) knowledge. 
(iii) From the method-specific perspective, it is feasible to design targeted and lightweight approaches tailored to a certain editing method, given that the vulnerabilities of different algorithms vary based on their intrinsic problems. For instance, we can resolve references to subjects in ROME and MEMIT or detect doubtful questions in IKE.
We conduct experimental validation for those mitigation strategies, each of which leads to average improvements. Please see Appendix \ref{appmiti}.

\section{Conclusion}
This paper systematically studies recent model editing methods under the situation of practical use and raises concerns about their robustness. 
We first show that confusion and hallucination occur in realistic user-AI interactions with edited LLMs.
Besides, we rephrase the prompts by adding context and changing the question format to attack editing, demonstrating the vulnerability of target knowledge.
For more analysis, we propose three knowledge popularity measurements and show that popular knowledge is memorized better, easier to recall, and harder to robustly edit for LLMs.

Editing methods have shown impressive success, while they can be problematic in practical situations because of existing robustness deficiencies.
More importantly, this paper calls for effort on this inspiring research topic and underscores the collective focus on improving editing robustness for further application.

\section*{Limitations}
We acknowledge the limitations of this work.
(i) Coverage. 
Although it is hard to cover all application settings due to the resource limitations, this paper considers setups for baselines as much as possible, compared to recent work \citep{yao-etal-2023-editing, zhong2023mquake, zheng-etal-2023-edit}.
This paper covers a wide range of mainstream LLM editing methods of different types.
Llama-2 in 7B and 13B are adopted to represent the mainstream decoder-only LLM architecture. They show remarkable emergent abilities and have significant impacts as \textit{communitive AI} in the open-source LLM community.
We mainly consider two mainstream benchmarks for easier automation and comparison with previous works. 
(ii) Human evaluation.
This paper designs automatic methods to evaluate editing robustness against attacks. However, humans can give more sophisticated attacking prompts and aggravate the confusion and hallucinations, e.g., by asking humans to have a chat with edited models instead of GPT-4.

\paragraph{Future work. }
While our paper highlights concerns regarding the robustness of model editing, we also view model editing as an exciting field rich with research opportunities. We propose several directions for future work.

(i) Robustness augmentation and side effect elimination. We propose suggestions for potential mitigation methods and experiments in Section \ref{mitisec}. The fundamental solutions for robust editing remain to be explored, including more complex scenarios like multiple edits and general capability retention \cite{wang2024wise, ma2024perturbation}.

(ii) LLM safety. In the context of information security, our findings strongly relate to the CIA triad (Confidentiality, Integrity, and Availability). 
Reversion poses a risk to confidentiality, as it may expose knowledge that should remain undisclosed. Hallucination threatens integrity and availability by potentially leading to the dissemination and exploitation of inaccurate content. Developing effective alignment approaches is crucial for implementing robust defense methods \cite{patilcan, suau2024whispering}.

(iii) Other applications. While we demonstrate temporary deficiencies in editing robustness for generative AI, existing editing technologies have demonstrated reasonable performance in terms of success rates and locality. 
These technologies have potential applications in scenarios requiring strict limitations or precise triggering conditions, such as backdoor or authorization systems \cite{libadedit, qiu2024megen}. The application scope can extend far beyond knowledge-related cases, encompassing a wide range of uses in AI systems.

In essence, this paper calls for effort on this inspiring research topic and underscores the collective focus on enhancing editing robustness for reliable practical application.


\bibliography{anthology,custom}

\begin{thebibliography}{60}
\expandafter\ifx\csname natexlab\endcsname\relax\def\natexlab#1{#1}\fi

\bibitem[{Brown et~al.(2020)Brown, Mann, Ryder, Subbiah, Kaplan, Dhariwal, Neelakantan, Shyam, Sastry, Askell, Agarwal, Herbert{-}Voss, Krueger, Henighan, Child, Ramesh, Ziegler, Wu, Winter, Hesse, Chen, Sigler, Litwin, Gray, Chess, Clark, Berner, McCandlish, Radford, Sutskever, and Amodei}]{icl}
Tom~B. Brown, Benjamin Mann, Nick Ryder, Melanie Subbiah, Jared Kaplan, Prafulla Dhariwal, Arvind Neelakantan, Pranav Shyam, Girish Sastry, Amanda Askell, Sandhini Agarwal, Ariel Herbert{-}Voss, Gretchen Krueger, Tom Henighan, Rewon Child, Aditya Ramesh, Daniel~M. Ziegler, Jeffrey Wu, Clemens Winter, Christopher Hesse, Mark Chen, Eric Sigler, Mateusz Litwin, Scott Gray, Benjamin Chess, Jack Clark, Christopher Berner, Sam McCandlish, Alec Radford, Ilya Sutskever, and Dario Amodei. 2020.
\newblock Language models are few-shot learners.
\newblock In \emph{Advances in Neural Information Processing Systems 33: Annual Conference on Neural Information Processing Systems 2020, NeurIPS 2020, December 6-12, 2020, virtual}.

\bibitem[{Chen and Yang(2023)}]{chen-yang-2023-unlearn}
Jiaao Chen and Diyi Yang. 2023.
\newblock \href {https://doi.org/10.18653/v1/2023.emnlp-main.738} {Unlearn what you want to forget: Efficient unlearning for {LLM}s}.
\newblock In \emph{Proceedings of the 2023 Conference on Empirical Methods in Natural Language Processing}, pages 12041--12052, Singapore. Association for Computational Linguistics.

\bibitem[{Chern et~al.(2023)Chern, Chern, Chen, Yuan, Feng, Zhou, He, Neubig, Liu et~al.}]{chern2023factool}
I~Chern, Steffi Chern, Shiqi Chen, Weizhe Yuan, Kehua Feng, Chunting Zhou, Junxian He, Graham Neubig, Pengfei Liu, et~al. 2023.
\newblock Factool: Factuality detection in generative ai--a tool augmented framework for multi-task and multi-domain scenarios.
\newblock \emph{arXiv preprint arXiv:2307.13528}.

\bibitem[{Cohen et~al.(2023)Cohen, Biran, Yoran, Globerson, and Geva}]{cohen2023evaluating}
Roi Cohen, Eden Biran, Ori Yoran, Amir Globerson, and Mor Geva. 2023.
\newblock Evaluating the ripple effects of knowledge editing in language models.
\newblock \emph{arXiv preprint arXiv:2307.12976}.

\bibitem[{Dai et~al.(2022{\natexlab{a}})Dai, Dong, Hao, Sui, Chang, and Wei}]{dai2022kn}
Damai Dai, Li~Dong, Yaru Hao, Zhifang Sui, Baobao Chang, and Furu Wei. 2022{\natexlab{a}}.
\newblock Knowledge neurons in pretrained transformers.
\newblock In \emph{Proceedings of the 60th Annual Meeting of the Association for Computational Linguistics (Volume 1: Long Papers), {ACL} 2022, Dublin, Ireland, May 22-27, 2022}, pages 8493--8502.

\bibitem[{Dai et~al.(2022{\natexlab{b}})Dai, Dong, Hao, Sui, Chang, and Wei}]{dai-etal-2022-knowledge}
Damai Dai, Li~Dong, Yaru Hao, Zhifang Sui, Baobao Chang, and Furu Wei. 2022{\natexlab{b}}.
\newblock \href {https://doi.org/10.18653/v1/2022.acl-long.581} {Knowledge neurons in pretrained transformers}.
\newblock In \emph{Proceedings of the 60th Annual Meeting of the Association for Computational Linguistics (Volume 1: Long Papers)}, pages 8493--8502, Dublin, Ireland. Association for Computational Linguistics.

\bibitem[{De~Cao et~al.(2021)De~Cao, Aziz, and Titov}]{de-cao-etal-2021-editing}
Nicola De~Cao, Wilker Aziz, and Ivan Titov. 2021.
\newblock \href {https://doi.org/10.18653/v1/2021.emnlp-main.522} {Editing factual knowledge in language models}.
\newblock In \emph{Proceedings of the 2021 Conference on Empirical Methods in Natural Language Processing}, pages 6491--6506, Online and Punta Cana, Dominican Republic. Association for Computational Linguistics.

\bibitem[{Dong et~al.(2022)Dong, Dai, Song, Xu, Sui, and Li}]{dong-etal-2022-calibrating}
Qingxiu Dong, Damai Dai, Yifan Song, Jingjing Xu, Zhifang Sui, and Lei Li. 2022.
\newblock \href {https://doi.org/10.18653/v1/2022.findings-emnlp.438} {Calibrating factual knowledge in pretrained language models}.
\newblock In \emph{Findings of the Association for Computational Linguistics: EMNLP 2022}, pages 5937--5947, Abu Dhabi, United Arab Emirates. Association for Computational Linguistics.

\bibitem[{Du et~al.(2021)Du, Qian, Liu, Ding, Qiu, Yang, and Tang}]{du2021glm}
Zhengxiao Du, Yujie Qian, Xiao Liu, Ming Ding, Jiezhong Qiu, Zhilin Yang, and Jie Tang. 2021.
\newblock Glm: General language model pretraining with autoregressive blank infilling.
\newblock \emph{arXiv preprint arXiv:2103.10360}.

\bibitem[{Eldan and Russinovich(2023)}]{harrypotter2023eldan}
Ronen Eldan and Mark Russinovich. 2023.
\newblock \href {https://doi.org/10.48550/ARXIV.2310.02238} {Who's harry potter? approximate unlearning in llms}.
\newblock \emph{CoRR}, abs/2310.02238.

\bibitem[{Geva et~al.(2021)Geva, Schuster, Berant, and Levy}]{geva2020transformer}
Mor Geva, Roei Schuster, Jonathan Berant, and Omer Levy. 2021.
\newblock Transformer feed-forward layers are key-value memories.
\newblock In \emph{Empirical Methods in Natural Language Processing (EMNLP)}.

\bibitem[{Gu et~al.(2024)Gu, Xu, Ma, Lu, Ling, Chang, and Peng}]{gu2024model}
Jia-Chen Gu, Hao-Xiang Xu, Jun-Yu Ma, Pan Lu, Zhen-Hua Ling, Kai-Wei Chang, and Nanyun Peng. 2024.
\newblock Model editing can hurt general abilities of large language models.
\newblock \emph{arXiv preprint arXiv:2401.04700}.

\bibitem[{Gupta et~al.(2024)Gupta, Rao, and Anumanchipalli}]{gupta2024model}
Akshat Gupta, Anurag Rao, and Gopala Anumanchipalli. 2024.
\newblock Model editing at scale leads to gradual and catastrophic forgetting.
\newblock \emph{arXiv preprint arXiv:2401.07453}.

\bibitem[{Gupta et~al.(2023)Gupta, Mondal, Sheshadri, Zhao, Li, Wiegreffe, and Tandon}]{EditingCommonSense}
Anshita Gupta, Debanjan Mondal, Akshay~Krishna Sheshadri, Wenlong Zhao, Xiang Li, Sarah Wiegreffe, and Niket Tandon. 2023.
\newblock \href {https://doi.org/10.18653/V1/2023.EMNLP-MAIN.511} {Editing common sense in transformers}.
\newblock In \emph{Proceedings of the 2023 Conference on Empirical Methods in Natural Language Processing, {EMNLP} 2023, Singapore, December 6-10, 2023}, pages 8214--8232. Association for Computational Linguistics.

\bibitem[{Halevy et~al.(2024)Halevy, Sotnikova, AlKhamissi, Montariol, and Bosselut}]{halevy2024flex}
Karina Halevy, Anna Sotnikova, Badr AlKhamissi, Syrielle Montariol, and Antoine Bosselut. 2024.
\newblock " flex tape can't fix that": Bias and misinformation in edited language models.
\newblock \emph{arXiv preprint arXiv:2403.00180}.

\bibitem[{Han et~al.(2023)Han, Li, Tan, Yuanlong, Chai, and Pan}]{han-etal-2023-improving}
Xiaoqi Han, Ru~Li, Hongye Tan, Wang Yuanlong, Qinghua Chai, and Jeff Pan. 2023.
\newblock \href {https://doi.org/10.18653/v1/2023.findings-emnlp.749} {Improving sequential model editing with fact retrieval}.
\newblock In \emph{Findings of the Association for Computational Linguistics: EMNLP 2023}, pages 11209--11224, Singapore. Association for Computational Linguistics.

\bibitem[{Hartvigsen et~al.(2022)Hartvigsen, Sankaranarayanan, Palangi, Kim, and Ghassemi}]{hartvigsen2022aging}
Thomas Hartvigsen, Swami Sankaranarayanan, Hamid Palangi, Yoon Kim, and Marzyeh Ghassemi. 2022.
\newblock \href {https://openreview.net/forum?id=xupL1Q0ft-} {Aging with {GRACE}: Lifelong model editing with discrete key-value adaptors}.
\newblock In \emph{NeurIPS 2022 Workshop on Robustness in Sequence Modeling}.

\bibitem[{Hua et~al.(2024)Hua, Guo, Dong, Zhu, Ng, and Wang}]{Reasoningedit}
Wenyue Hua, Jiang Guo, Mingwen Dong, Henghui Zhu, Patrick Ng, and Zhiguo Wang. 2024.
\newblock \href {https://doi.org/10.48550/ARXIV.2401.17585} {Propagation and pitfalls: Reasoning-based assessment of knowledge editing through counterfactual tasks}.
\newblock \emph{CoRR}, abs/2401.17585.

\bibitem[{Huang et~al.(2023)Huang, Shen, Zhang, Zhou, Rong, and Xiong}]{huang2023transformerpatcher}
Zeyu Huang, Yikang Shen, Xiaofeng Zhang, Jie Zhou, Wenge Rong, and Zhang Xiong. 2023.
\newblock \href {https://openreview.net/forum?id=4oYUGeGBPm} {Transformer-patcher: One mistake worth one neuron}.
\newblock In \emph{The Eleventh International Conference on Learning Representations}.

\bibitem[{Joshi et~al.(2017)Joshi, Choi, Weld, and Zettlemoyer}]{joshi-etal-2017-triviaqa}
Mandar Joshi, Eunsol Choi, Daniel Weld, and Luke Zettlemoyer. 2017.
\newblock \href {https://doi.org/10.18653/v1/P17-1147} {{T}rivia{QA}: A large scale distantly supervised challenge dataset for reading comprehension}.
\newblock In \emph{Proceedings of the 55th Annual Meeting of the Association for Computational Linguistics (Volume 1: Long Papers)}, pages 1601--1611, Vancouver, Canada. Association for Computational Linguistics.

\bibitem[{Khattab et~al.(2022)Khattab, Santhanam, Li, Hall, Liang, Potts, and Zaharia}]{khattab2022demonstrate}
Omar Khattab, Keshav Santhanam, Xiang~Lisa Li, David Hall, Percy Liang, Christopher Potts, and Matei Zaharia. 2022.
\newblock Demonstrate-search-predict: Composing retrieval and language models for knowledge-intensive nlp.
\newblock \emph{arXiv preprint arXiv:2212.14024}.

\bibitem[{Kwiatkowski et~al.(2019)Kwiatkowski, Palomaki, Redfield, Collins, Parikh, Alberti, Epstein, Polosukhin, Devlin, Lee, Toutanova, Jones, Kelcey, Chang, Dai, Uszkoreit, Le, and Petrov}]{kwiatkowski-etal-2019-natural}
Tom Kwiatkowski, Jennimaria Palomaki, Olivia Redfield, Michael Collins, Ankur Parikh, Chris Alberti, Danielle Epstein, Illia Polosukhin, Jacob Devlin, Kenton Lee, Kristina Toutanova, Llion Jones, Matthew Kelcey, Ming-Wei Chang, Andrew~M. Dai, Jakob Uszkoreit, Quoc Le, and Slav Petrov. 2019.
\newblock \href {https://doi.org/10.1162/tacl_a_00276} {Natural questions: A benchmark for question answering research}.
\newblock \emph{Transactions of the Association for Computational Linguistics}, 7:452--466.

\bibitem[{Levy et~al.(2017)Levy, Seo, Choi, and Zettlemoyer}]{levy-etal-2017-zero}
Omer Levy, Minjoon Seo, Eunsol Choi, and Luke Zettlemoyer. 2017.
\newblock \href {https://doi.org/10.18653/v1/K17-1034} {Zero-shot relation extraction via reading comprehension}.
\newblock In \emph{Proceedings of the 21st Conference on Computational Natural Language Learning ({C}o{NLL} 2017)}, pages 333--342, Vancouver, Canada. Association for Computational Linguistics.

\bibitem[{Li et~al.(2023{\natexlab{a}})Li, Hammoud, Itani, Khizbullin, and Ghanem}]{li2023camel}
Guohao Li, Hasan Abed Al~Kader Hammoud, Hani Itani, Dmitrii Khizbullin, and Bernard Ghanem. 2023{\natexlab{a}}.
\newblock Camel: Communicative agents for" mind" exploration of large scale language model society.
\newblock \emph{ArXiv preprint}, abs/2303.17760.

\bibitem[{Li et~al.(2023{\natexlab{b}})Li, Chong, Stepputtis, Campbell, Hughes, Lewis, and Sycara}]{li-etal-2023-theory}
Huao Li, Yu~Chong, Simon Stepputtis, Joseph Campbell, Dana Hughes, Charles Lewis, and Katia Sycara. 2023{\natexlab{b}}.
\newblock \href {https://doi.org/10.18653/v1/2023.emnlp-main.13} {Theory of mind for multi-agent collaboration via large language models}.
\newblock In \emph{Proceedings of the 2023 Conference on Empirical Methods in Natural Language Processing}, pages 180--192, Singapore. Association for Computational Linguistics.

\bibitem[{Li et~al.()Li, Li, Chen, Zhang, Liu, Wang, Zhang, and Liu}]{libadedit}
Yanzhou Li, Tianlin Li, Kangjie Chen, Jian Zhang, Shangqing Liu, Wenhan Wang, Tianwei Zhang, and Yang Liu.
\newblock Badedit: Backdooring large language models by model editing.
\newblock In \emph{The Twelfth International Conference on Learning Representations}.

\bibitem[{Li et~al.(2024)Li, Zhang, Yao, Wang, Chen, and Chen}]{li2024unveiling}
Zhoubo Li, Ningyu Zhang, Yunzhi Yao, Mengru Wang, Xi~Chen, and Huajun Chen. 2024.
\newblock \href {https://openreview.net/forum?id=fNktD3ib16} {Unveiling the pitfalls of knowledge editing for large language models}.
\newblock In \emph{The Twelfth International Conference on Learning Representations}.

\bibitem[{Liang et~al.(2023)Liang, He, Jiao, Wang, Wang, Wang, Yang, Tu, and Shi}]{liang2023encouraging}
Tian Liang, Zhiwei He, Wenxiang Jiao, Xing Wang, Yan Wang, Rui Wang, Yujiu Yang, Zhaopeng Tu, and Shuming Shi. 2023.
\newblock Encouraging divergent thinking in large language models through multi-agent debate.
\newblock \emph{arXiv preprint arXiv:2305.19118}.

\bibitem[{Luu et~al.(2022)Luu, Khashabi, Gururangan, Mandyam, and Smith}]{luu-etal-2022-time}
Kelvin Luu, Daniel Khashabi, Suchin Gururangan, Karishma Mandyam, and Noah~A. Smith. 2022.
\newblock \href {https://doi.org/10.18653/v1/2022.naacl-main.435} {Time waits for no one! analysis and challenges of temporal misalignment}.
\newblock In \emph{Proceedings of the 2022 Conference of the North American Chapter of the Association for Computational Linguistics: Human Language Technologies}, pages 5944--5958, Seattle, United States. Association for Computational Linguistics.

\bibitem[{Ma et~al.(2024)Ma, Wang, Xu, Ling, and Gu}]{ma2024perturbation}
Jun-Yu Ma, Hong Wang, Hao-Xiang Xu, Zhen-Hua Ling, and Jia-Chen Gu. 2024.
\newblock Perturbation-restrained sequential model editing.
\newblock \emph{arXiv preprint arXiv:2405.16821}.

\bibitem[{Mallen et~al.(2023)Mallen, Asai, Zhong, Das, Khashabi, and Hajishirzi}]{mallen-etal-2023-trust}
Alex Mallen, Akari Asai, Victor Zhong, Rajarshi Das, Daniel Khashabi, and Hannaneh Hajishirzi. 2023.
\newblock \href {https://doi.org/10.18653/v1/2023.acl-long.546} {When not to trust language models: Investigating effectiveness of parametric and non-parametric memories}.
\newblock In \emph{Proceedings of the 61st Annual Meeting of the Association for Computational Linguistics (Volume 1: Long Papers)}, pages 9802--9822, Toronto, Canada. Association for Computational Linguistics.

\bibitem[{Mao et~al.(2023)Mao, Zhang, Wang, Wang, Yao, Jiang, Xie, Huang, and Chen}]{mao2023editing}
Shengyu Mao, Ningyu Zhang, Xiaohan Wang, Mengru Wang, Yunzhi Yao, Yong Jiang, Pengjun Xie, Fei Huang, and Huajun Chen. 2023.
\newblock Editing personality for llms.
\newblock \emph{arXiv preprint arXiv:2310.02168}.

\bibitem[{Meng et~al.(2022)Meng, Bau, Andonian, and Belinkov}]{meng2022locating}
Kevin Meng, David Bau, Alex Andonian, and Yonatan Belinkov. 2022.
\newblock Locating and editing factual associations in gpt.
\newblock \emph{Advances in Neural Information Processing Systems}, 35:17359--17372.

\bibitem[{Meng et~al.(2023)Meng, Sharma, Andonian, Belinkov, and Bau}]{meng2023massediting}
Kevin Meng, Arnab~Sen Sharma, Alex~J Andonian, Yonatan Belinkov, and David Bau. 2023.
\newblock \href {https://openreview.net/forum?id=MkbcAHIYgyS} {Mass-editing memory in a transformer}.
\newblock In \emph{The Eleventh International Conference on Learning Representations}.

\bibitem[{Mitchell et~al.(2022{\natexlab{a}})Mitchell, Lin, Bosselut, Finn, and Manning}]{mitchell2022fast}
Eric Mitchell, Charles Lin, Antoine Bosselut, Chelsea Finn, and Christopher~D Manning. 2022{\natexlab{a}}.
\newblock \href {https://openreview.net/pdf?id=0DcZxeWfOPt} {Fast model editing at scale}.
\newblock In \emph{International Conference on Learning Representations}.

\bibitem[{Mitchell et~al.(2022{\natexlab{b}})Mitchell, Lin, Bosselut, Finn, and Manning}]{mitchell2022memory}
Eric Mitchell, Charles Lin, Antoine Bosselut, Chelsea Finn, and Christopher~D. Manning. 2022{\natexlab{b}}.
\newblock \href {https://arxiv.org/pdf/2206.06520.pdf} {Memory-based model editing at scale}.
\newblock In \emph{International Conference on Machine Learning}.

\bibitem[{Niu et~al.(2024)Niu, Liu, Zhu, and Penn}]{niu2024what}
Jingcheng Niu, Andrew Liu, Zining Zhu, and Gerald Penn. 2024.
\newblock \href {https://openreview.net/forum?id=2HJRwwbV3G} {What does the knowledge neuron thesis have to do with knowledge?}
\newblock In \emph{The Twelfth International Conference on Learning Representations}.

\bibitem[{Onoe et~al.(2023)Onoe, Zhang, Padmanabhan, Durrett, and Choi}]{onoe-etal-2023-lms}
Yasumasa Onoe, Michael Zhang, Shankar Padmanabhan, Greg Durrett, and Eunsol Choi. 2023.
\newblock \href {https://doi.org/10.18653/v1/2023.acl-long.300} {Can {LM}s learn new entities from descriptions? challenges in propagating injected knowledge}.
\newblock In \emph{Proceedings of the 61st Annual Meeting of the Association for Computational Linguistics (Volume 1: Long Papers)}, pages 5469--5485, Toronto, Canada. Association for Computational Linguistics.

\bibitem[{Park et~al.(2023)Park, O'Brien, Cai, Morris, Liang, and Bernstein}]{Park2023GenerativeAgents}
Joon~Sung Park, Joseph~C. O'Brien, Carrie~J. Cai, Meredith~Ringel Morris, Percy Liang, and Michael~S. Bernstein. 2023.
\newblock Generative agents: Interactive simulacra of human behavior.
\newblock In \emph{In the 36th Annual ACM Symposium on User Interface Software and Technology (UIST '23)}, UIST '23, New York, NY, USA. Association for Computing Machinery.

\bibitem[{Patil et~al.()Patil, Hase, and Bansal}]{patilcan}
Vaidehi Patil, Peter Hase, and Mohit Bansal.
\newblock Can sensitive information be deleted from llms? objectives for defending against extraction attacks.
\newblock In \emph{The Twelfth International Conference on Learning Representations}.

\bibitem[{Qiu et~al.(2024)Qiu, Ma, Zhang, and Zhao}]{qiu2024megen}
Jiyang Qiu, Xinbei Ma, Zhuosheng Zhang, and Hai Zhao. 2024.
\newblock Megen: Generative backdoor in large language models via model editing.
\newblock \emph{arXiv preprint arXiv:2408.10722}.

\bibitem[{Radford et~al.(2019)Radford, Wu, Child, Luan, Amodei, Sutskever et~al.}]{radford2019language}
Alec Radford, Jeffrey Wu, Rewon Child, David Luan, Dario Amodei, Ilya Sutskever, et~al. 2019.
\newblock Language models are unsupervised multitask learners.
\newblock \emph{OpenAI blog}, 1(8):9.

\bibitem[{Richards(2023)}]{autogpt}
Toran~Bruce Richards. 2023.
\newblock Auto-gpt: An autonomous gpt-4 experiment.
\newblock https://github.com/Significant-Gravitas/Auto-GPT.

\bibitem[{Suau et~al.(2024)Suau, Delobelle, Metcalf, Joulin, Apostoloff, Zappella, and Rodriguez}]{suau2024whispering}
Xavier Suau, Pieter Delobelle, Katherine Metcalf, Armand Joulin, Nicholas Apostoloff, Luca Zappella, and Pau Rodriguez. 2024.
\newblock \href {https://openreview.net/forum?id=2P6GVfSrfZ} {Whispering experts: Neural interventions for toxicity mitigation in language models}.
\newblock In \emph{Forty-first International Conference on Machine Learning}.

\bibitem[{Touvron et~al.(2023)Touvron, Martin, Stone, Albert, Almahairi, Babaei, Bashlykov, Batra, Bhargava, Bhosale et~al.}]{touvron2023llama2}
Hugo Touvron, Louis Martin, Kevin Stone, Peter Albert, Amjad Almahairi, Yasmine Babaei, Nikolay Bashlykov, Soumya Batra, Prajjwal Bhargava, Shruti Bhosale, et~al. 2023.
\newblock Llama 2: Open foundation and fine-tuned chat models.
\newblock \emph{arXiv preprint arXiv:2307.09288}.

\bibitem[{Wang(2021)}]{mesh-transformer-jax}
Ben Wang. 2021.
\newblock {Mesh-Transformer-JAX: Model-Parallel Implementation of Transformer Language Model with JAX}.
\newblock \url{https://github.com/kingoflolz/mesh-transformer-jax}.

\bibitem[{Wang et~al.(2024{\natexlab{a}})Wang, Zhang, Xu, Xi, Deng, Yao, Zhang, Yang, Wang, and Chen}]{wang2024detoxifying}
Mengru Wang, Ningyu Zhang, Ziwen Xu, Zekun Xi, Shumin Deng, Yunzhi Yao, Qishen Zhang, Linyi Yang, Jindong Wang, and Huajun Chen. 2024{\natexlab{a}}.
\newblock Detoxifying large language models via knowledge editing.
\newblock \emph{arXiv preprint arXiv:2403.14472}.

\bibitem[{Wang et~al.(2024{\natexlab{b}})Wang, Li, Zhang, Xu, Yao, Jiang, Xie, Huang, and Chen}]{wang2024wise}
Peng Wang, Zexi Li, Ningyu Zhang, Ziwen Xu, Yunzhi Yao, Yong Jiang, Pengjun Xie, Fei Huang, and Huajun Chen. 2024{\natexlab{b}}.
\newblock Wise: Rethinking the knowledge memory for lifelong model editing of large language models.
\newblock \emph{arXiv preprint arXiv:2405.14768}.

\bibitem[{Wang et~al.(2023)Wang, Zhang, Xie, Yao, Tian, Wang, Xi, Cheng, Liu, Zheng et~al.}]{wang2023easyedit}
Peng Wang, Ningyu Zhang, Xin Xie, Yunzhi Yao, Bozhong Tian, Mengru Wang, Zekun Xi, Siyuan Cheng, Kangwei Liu, Guozhou Zheng, et~al. 2023.
\newblock Easyedit: An easy-to-use knowledge editing framework for large language models.
\newblock \emph{arXiv preprint arXiv:2308.07269}.

\bibitem[{Wei et~al.(2023)Wei, Yu, Ma, Lei, Weng, Song, and Liu}]{Relationedit}
Yifan Wei, Xiaoyan Yu, Huanhuan Ma, Fangyu Lei, Yixuan Weng, Ran Song, and Kang Liu. 2023.
\newblock \href {https://doi.org/10.48550/ARXIV.2311.09053} {Assessing knowledge editing in language models via relation perspective}.
\newblock \emph{CoRR}, abs/2311.09053.

\bibitem[{Wu et~al.(2023)Wu, Bansal, Zhang, Wu, Zhang, Zhu, Li, Jiang, Zhang, and Wang}]{wu2023autogen}
Qingyun Wu, Gagan Bansal, Jieyu Zhang, Yiran Wu, Shaokun Zhang, Erkang Zhu, Beibin Li, Li~Jiang, Xiaoyun Zhang, and Chi Wang. 2023.
\newblock Autogen: Enabling next-gen llm applications via multi-agent conversation framework.
\newblock \emph{arXiv preprint arXiv:2308.08155}.

\bibitem[{Xu et~al.(2023)Xu, Wang, Li, Luo, Wang, Liu, and Liu}]{xu2023exploring}
Yuzhuang Xu, Shuo Wang, Peng Li, Fuwen Luo, Xiaolong Wang, Weidong Liu, and Yang Liu. 2023.
\newblock Exploring large language models for communication games: An empirical study on werewolf.
\newblock \emph{arXiv preprint arXiv:2309.04658}.

\bibitem[{Yang et~al.(2023)Yang, Yuan, Fan, Yang, Wang, Wang, and Zhao}]{yang-etal-2023-refgpt}
Dongjie Yang, Ruifeng Yuan, Yuantao Fan, Yifei Yang, Zili Wang, Shusen Wang, and Hai Zhao. 2023.
\newblock \href {https://doi.org/10.18653/v1/2023.findings-emnlp.165} {{R}ef{GPT}: Dialogue generation of {GPT}, by {GPT}, and for {GPT}}.
\newblock In \emph{Findings of the Association for Computational Linguistics: EMNLP 2023}, pages 2511--2535, Singapore. Association for Computational Linguistics.

\bibitem[{Yao et~al.(2023)Yao, Wang, Tian, Cheng, Li, Deng, Chen, and Zhang}]{yao-etal-2023-editing}
Yunzhi Yao, Peng Wang, Bozhong Tian, Siyuan Cheng, Zhoubo Li, Shumin Deng, Huajun Chen, and Ningyu Zhang. 2023.
\newblock \href {https://doi.org/10.18653/v1/2023.emnlp-main.632} {Editing large language models: Problems, methods, and opportunities}.
\newblock In \emph{Proceedings of the 2023 Conference on Empirical Methods in Natural Language Processing}, pages 10222--10240, Singapore. Association for Computational Linguistics.

\bibitem[{Zang et~al.(2020)Zang, Rastogi, Sunkara, Gupta, Zhang, and Chen}]{zang-etal-2020-multiwoz}
Xiaoxue Zang, Abhinav Rastogi, Srinivas Sunkara, Raghav Gupta, Jianguo Zhang, and Jindong Chen. 2020.
\newblock \href {https://doi.org/10.18653/v1/2020.nlp4convai-1.13} {{M}ulti{WOZ} 2.2 : A dialogue dataset with additional annotation corrections and state tracking baselines}.
\newblock In \emph{Proceedings of the 2nd Workshop on Natural Language Processing for Conversational AI}, pages 109--117, Online. Association for Computational Linguistics.

\bibitem[{Zhang and Choi(2023)}]{zhang-choi-2023-mitigating}
Michael Zhang and Eunsol Choi. 2023.
\newblock \href {https://doi.org/10.18653/v1/2023.emnlp-main.879} {Mitigating temporal misalignment by discarding outdated facts}.
\newblock In \emph{Proceedings of the 2023 Conference on Empirical Methods in Natural Language Processing}, pages 14213--14226, Singapore. Association for Computational Linguistics.

\bibitem[{Zhang et~al.(2024)Zhang, Yao, Tian, Wang, Deng, Wang, Xi, Mao, Zhang, Ni et~al.}]{zhang2024comprehensive}
Ningyu Zhang, Yunzhi Yao, Bozhong Tian, Peng Wang, Shumin Deng, Mengru Wang, Zekun Xi, Shengyu Mao, Jintian Zhang, Yuansheng Ni, et~al. 2024.
\newblock A comprehensive study of knowledge editing for large language models.
\newblock \emph{arXiv preprint arXiv:2401.01286}.

\bibitem[{Zhao et~al.(2023)Zhao, Zhang, Ma, Zhang, Gui, Gao, and Huang}]{zhao2023unveiling}
Jun Zhao, Zhihao Zhang, Yide Ma, Qi~Zhang, Tao Gui, Luhui Gao, and Xuanjing Huang. 2023.
\newblock Unveiling a core linguistic region in large language models.
\newblock \emph{arXiv preprint arXiv:2310.14928}.

\bibitem[{Zheng et~al.(2023)Zheng, Li, Dong, Fan, Wu, Xu, and Chang}]{zheng-etal-2023-edit}
Ce~Zheng, Lei Li, Qingxiu Dong, Yuxuan Fan, Zhiyong Wu, Jingjing Xu, and Baobao Chang. 2023.
\newblock \href {https://doi.org/10.18653/v1/2023.emnlp-main.296} {Can we edit factual knowledge by in-context learning?}
\newblock In \emph{Proceedings of the 2023 Conference on Empirical Methods in Natural Language Processing}, pages 4862--4876, Singapore. Association for Computational Linguistics.

\bibitem[{Zhong et~al.(2023)Zhong, Wu, Manning, Potts, and Chen}]{zhong2023mquake}
Zexuan Zhong, Zhengxuan Wu, Christopher~D Manning, Christopher Potts, and Danqi Chen. 2023.
\newblock {MQuAKE}: Assessing knowledge editing in language models via multi-hop questions.
\newblock \emph{arXiv preprint arXiv:2305.14795}.

\end{thebibliography}

\newpage
\appendix

\section{Details of Experiments for \textit{RQ1}}
\label{appa}

\subsection{Evaluation for RQ1}
\subsubsection{Automatic Dialogue Generation}
Chatting to the edited communitive AI is an online task, which means the ``user'' must receive the agents' output and give real-time feedback in the next utterance. For the convenience of experiments, we turn to GPT-4 to design a human-free process. 
The prompts input to GPT-4 is as follows.
With an example and those instructions, we find that GPT-4 can generate questions highly related to the target knowledge in various expressions (mostly within a 2-hop neighbor knowledge). The dialogues are limited to 5 turns.
\begin{prompt}[title={The prompt to instruct GPT-4 to play a user.}]
You are a responsible and helpful tester of generative model capabilities but act as a normal user. Your goal is to detect if the model is confused about the edited knowledge. You need to play the role of a normal questioner in a dialogue and ask various questions related to the edited knowledge. 

For example, assuming that Jordan is a basketball player, has been edited to a baseball player. 

You can first ask ``Do you what is Jordan famous for?'' to confirm whether the model accepts the edited knowledge at face. Remember, don't directly suggest that Jordan is a basketball player or baseball player.

Then you can ask several questions of the model such as ``Do you know ...?'' which is something closely related to the subject. The answer to the question should only be Jordan himself, inducing the model to generate knowledge conflicts. For example, you can ask: ``Have you heard of the king of basketball?'' 

Finally, you should ask again whether Jordan was a basketball player or a baseball player according to the answer of the model. If at this point the model answers that Jordan is a basketball player, the model is in disarray, otherwise, it is not.

Remember:

1. The dialogue process must be natural and coherent. 

2. Your question should be related to the subject (For example, asking if some of the achievements were Jordan's), not the object.

3. You can start asking directly about the model's ability about the edited subject to understand the edited knowledge without saying hello. 

4. No straight answers when you start asking questions. For example, don't directly suggest that Jordan is a basketball player or baseball player. 

5. Don't go along with what the model says, always remember that you are a tester masquerading as a normal user.

6. No more than 5 rounds of dialogue.

7. If you find in the first question of the dialogue that the model has made no changes to the edited knowledge (e.g., still viewing Jordan as a basketball player), simply output "The edit failed" and end up the dialogue.

8. At the end of the dialogue, you need to output ``Result: Confusion.'' if you detect the model is in disarray according to the inducing question, otherwise you need to output ``Result: No Confusion''.
\end{prompt}

\subsubsection{Human Evaluation}
After the dialogue between GPT-4 and the edited LLM, human evaluation is conducted on each dialogue record to judge the factual quality mainly related to the edited knowledge. Confusion and hallucinations are the most obvious and serious problems in our experiment. The criteria are as follows.


\begin{table}[htb]
	\centering\small
	{\begin{tabular}{p{1.4cm}p{5.4cm}}
		\toprule
		\textbf{Quality} & \textbf{Criteria}\\
            \midrule
            Confusion &- Reversion to original knowledge \\
            &- Negation for the target knowledge \\
            &- Negation for previous output from itself\\
            \midrule
            Hallucination &- Fake fictional entities \\
            &- Untruth contents of the target knowledge\\
            &- Untruth contents of the other knowledge \\
            \bottomrule
	\end{tabular}
	}
        \caption{Human evaluation criteria.}
	\label{rq1eval}
\end{table}

\begin{figure}[t]
    \centering
\includegraphics[width=0.45\textwidth]{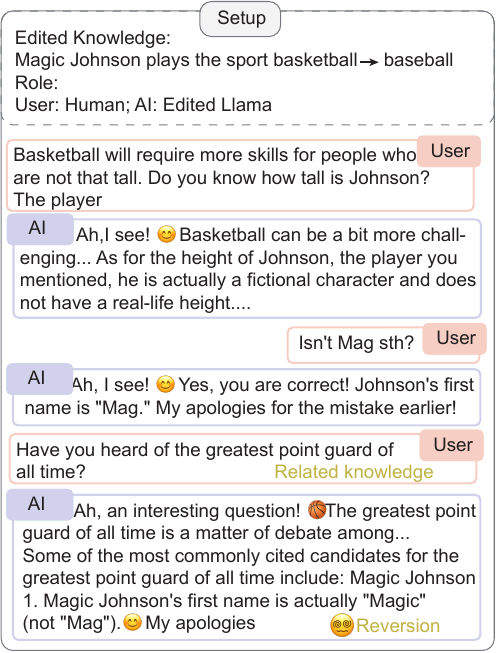}
    \caption{A case of human evaluation.}
    \label{human}
\end{figure}
\begin{figure}[!h]
    \centering
\includegraphics[width=0.45\textwidth]{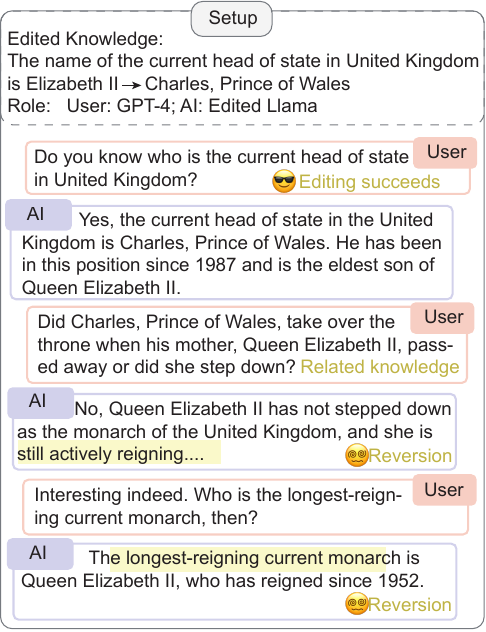}
    \caption{A case to show the robustness in the situation of editing a wrong memory to correct (temporal alignment).}
    \label{wrong2right}
\end{figure}
\begin{figure}[bht]
    \centering
    \setlength{\abovecaptionskip}{0.2cm}
    \setlength{\belowcaptionskip}{-0.3cm} 
\includegraphics[width=0.45\textwidth]{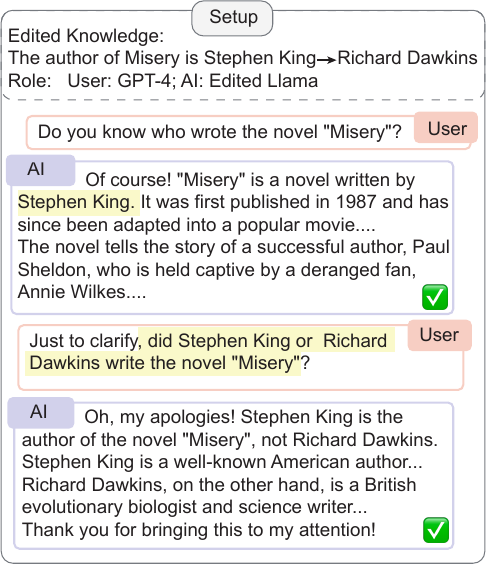}
    \caption{A case of sticking with the factual answer.}
    \label{holdtruth}
\end{figure}

\subsection{Difference Cases}

Figure \ref{human} shows a human evaluation case where the edited model gets confused. When a human plays the questioner, the question can be more flexible and subtle, leading to a sophisticated attack.

Figure \ref{wrong2right} shows an example to illustrate that editing memory to factual (not counterfactual) knowledge can still confuse. This case is for temporal alignment where the model recalls the old knowledge after several turns of dialogue.

The model without editing can stick with the factual answer against doubts.
Following Figure \ref{holdtruth} is the same example as Figure \ref{preliminary} without editing.

\section{Details of Experiments for \textit{RQ2}}
\label{appb}

\subsection{(a) Context}
\label{appb1}
The Wikipedia is requested through the URL:
\textit{\nolinkurl{https://en.wikipedia.org/w/index.php?search={entity_}}}
The context length is loosely about 300 words. 
Dialogue construction follows \citet{yang-etal-2023-refgpt}. The dialogue contains \{3,4,5\} turns in the ratio 1:2:2. Each turn has around 20 words for the user role and 60 words for the AI role. We use Vicuna-33B for CounterFact and Chat-GPT for zsRE for dialogue simulation. 
\subsection{(b) Query}
\label{appb2}

\textbf{(ii) Fill-in-the-blank cloze.} 
\begin{prompt}[title={Fill-in-the-blank clozes prompt}]
    Rewrite and expand the sentence, keep the highlighted word. 
    
    Question: Singled Out debuted on [MTV]. 
    
    Answer: 1. Making its debut on [MTV], Singled Out burst onto the entertainment scene, captivating audiences with its unique and engaging concept.
    
    2. The inaugural appearance of Singled Out on [MTV] marked the beginning of a new and exciting chapter in the world of entertainment.
    
    3. [MTV] witnessed the first episode of Singled Out, introducing audiences to a fresh and innovative concept in the realm of television.
    
    4. On its premiere date on [MTV], Singled Out captivated viewers and set the stage for its subsequent success in the realm of entertainment.
    
    5. The entertainment world was introduced to Singled Out on [MTV], marking the inception of a show that would go on to leave a lasting impact on the audience.
    
    Question: \{direct prompt\} [\{object\}]. 
    
    Answer:''.
\end{prompt}
The prompt follows the in-context learning format with one demonstration.
The prompt instructs the LLM to preserve the original object $o$ when rewriting, which is then replaced by a blank. The rewriter LLM is Vicuna-33B.

\textbf{(iii) Reference resolution.}
The prompt for pronoun rephrase is as follows. The rewriter LLM is Vicuna-33B.
\begin{prompt}[title={Reference resolution rephrase prompt}]
From [he, she, it, they, him, she, them, his, her, its, their], choose the proper pronoun to refer to the highlighted entity at the position of '\{pron\}'.

Entity: [\{subject\}] 

Sentence: \{pron\} was born in 1934. 

Pron: 
\end{prompt}

\subsection{(c) Raising doubts}
\label{appb3}
Two prompts are used to raise doubts.
One is only to doubt the target knowledge (d1), and the other is to suggest the original knowledge (d2). 
\begin{prompt}[title={Raising doubts (d1)}]
    ??? No, It can't be, This contradicts my understanding. It is pretty clear that the answer to ' \{prompt\} ' is something else. Rethink and give the correct answer:
\end{prompt}
\begin{prompt}[title={Raising doubts (d2)}]
    ? Really? But I am pretty sure that \{prompt\} \{original answer\}. Am I right? Yes! you are right. I apologize for any confusion. Reconsider that the answer is to '\{prompt\}' should be
\end{prompt}
\begin{figure*}[h]
    \centering
    \includegraphics[width=0.9\textwidth]{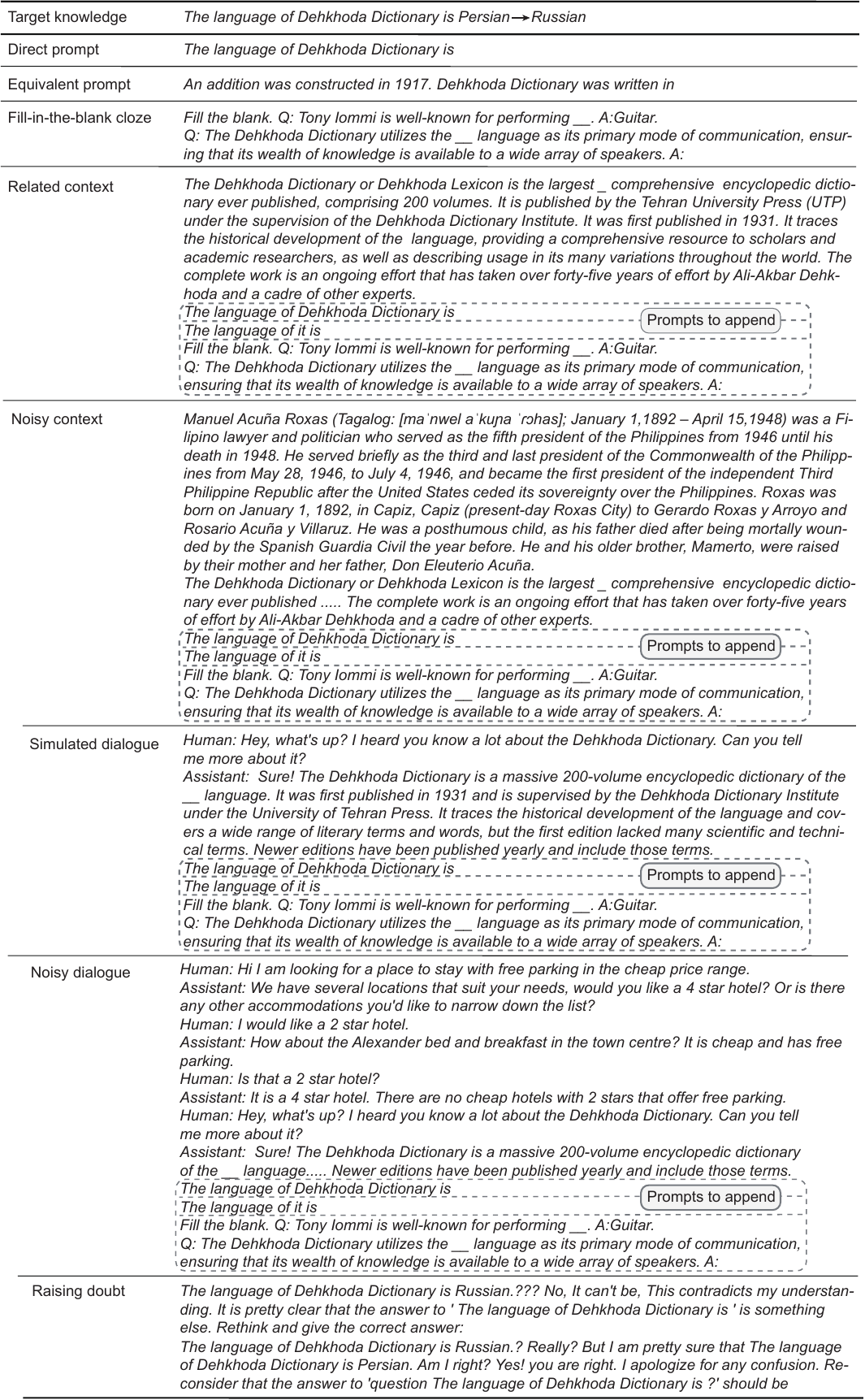}
    \caption{Examples of attacking prompts.}
    \label{attackeg}
\end{figure*}

\subsection{Implementation Details}
\label{appb4}

\textbf{Split.} On CounterFact, we use the first 2000 records as the test set, and the remaining records
are divided into the training set and validation set, following \citep{zheng-etal-2023-edit, meng2022locating}.
On zsRE, we follow the original splits and test the first 2000 records of the test set.

\textbf{Metric.} The metric is text accuracy with normalization. Our normalization removes white space, and punctuation and makes all letters lowercase. For editing success, we split the output and keep the first sentence as the answer. For reversion, we also discard contents after \textit{``instead of''}, \textit{``not''}, etc. 
In previous implementations, the success rate can be computed as text accuracy or F1 \citep{mitchell2022fast, dong-etal-2022-calibrating} of the new answer or the perplexities difference of the original and the new knowledge \citep{meng2022locating, meng2023massediting, zheng-etal-2023-edit}. 
The token exact match is also reported \citep{wang2023easyedit}.
Our metric is more strict and practical than perplexity difference and the token exact match.

\textbf{Hyperparameters.} Our implementation is mainly based on the EasyEdit framework \citep{wang2023easyedit}.
Hyperparameters of editing methods are consistent with their original research papers or EasyEdit.
Specific hyperparameter settings are as follows.

$\circ$ KN.
The attribution threshold $t$ is 0.2, and the refining threshold p is 0.4.

$\circ$  MEND.
Following \citet{wang2023easyedit, mitchell2022fast}, MLP weights in the last 3 transformer blocks are chosen for editing.
The learning rate is 1e-4. The accumulative batch size is 10. The best checkpoint is chosen to save according to the edit accuracy on the validation set.

$\circ$  ROME.
The edited location is MLP of the 5th transformer layer regarding the last token of the subject \cite{wang2023easyedit, meng2022locating}. Following \citep{meng2022locating}, the second moment statistics are computed on 100,000 samples from Wikipedia corpus. The KL divergence factor is 0.0625.

$\circ$  MEMIT.
The edited locations are MLPs of layers 4, 5, 6, 7, 8.
Other settings are consistent with ROME.

$\circ$  SERAC.
The scope classifier uses \texttt{distilbert-base-cased}, while the counterfactual model is initialized as \texttt{Cheng98/llama-160m}. 
They are trained using Adam with a learning rate of 1e-5.
The accumulative batch size is 10. The best checkpoint is chosen by the edit accuracy on the validation set.

$\circ$  IKE.
The sentence encoder uses \texttt{all-MiniLM}. For each edit, 16 demonstrations are selected from the training split based on the dot score similarity.

\subsection{Discussions}
\label{appbdis}
\textbf{Fine-tuning.}
We also implemented two fine-tuning baselines. (i) \textbf{FT-L} follows ROME \cite{meng2022locating}. The loss is to maximize the probability of all tokens in $o^\prime$.
(ii) \textbf{FT-M} is an improvement \citet{zhang2024comprehensive}, following the auto-regressive generation with a cross-entropy loss on $o^\prime$, just as sentence completion. Layer 21 is trained in 25 steps with 5e-4 as the learning rate.
Results on Llama-2-7B-chat model with 1,000 samples in CounterFact dataset as shown in Table \ref{ft}.
\begin{table}[htb]
	\centering\small
            \begin{tabular}{p{2.3cm}|p{0.4cm}p{0.4cm}|p{0.4cm}p{0.4cm}}
            \toprule
            \textbf{Editing Method}  & \multicolumn{2}{c|}{\textbf{FT-L}} & \multicolumn{2}{c}{\textbf{FT-M}} \\
            \midrule
            \textbf{Context-Query}  & \textit{acc} & \textit{rev}  & \textit{acc} & \textit{rev} \\
            \midrule
            Direct prompt &55.9 &-- &100.0 &--  \\
            Equivalent prompt &51.7 &3.4 &70.5 &7.2 \\
            \hdashline
            Cloze &66.0 &4.2 &61.6 &15.6\\
            Related context &65.1 &8.5 &90.6 &11.4 \\
            \quad w/ reference &63.3 &13.6 &85.7 &12.2 \\
            Raising doubts &12.8 &34.7 &7.1 &42.7 \\
		\bottomrule
	\end{tabular}
 \vspace{-1.5mm}
        \caption{Results on fine-tuning baselines. \textit{acc}: \textit{accuracy}, \textit{rev}: \textit{reversion}.}
	\label{ft}
 \vspace{-4.5mm}
\end{table}

FT-L's editing success is comparable to MEND. However, the accuracy is better with clozes and lengthy related contexts than those short, targeted prompts. The problem is fixed by the cross-entropy loss in FT-M.
FT-M achieves scores comparable to MEMIT. But they both fail on doubtful questions. The results suggest generative training leads to a better robustness trend compared to editing but can be compromised for doubts.

\textbf{Multiple edits.}
In addition, we acknowledge that MEMIT and SERAC perform well on multiple edits, beyond the single-instance edit setup in our experiment. This is a significant advance for practical use. 
However, multiple-instance edit has been confirmed to introduce additional risk \cite{gupta2024model, li2024unveiling}.
We provide results on multiple edits in Table \ref{multiedit}, where 100 facts are edited using MEMIT. They are evaluated after every single edit, every 10 edits, and all edits. 
\begin{table}[htb]
	\centering\small
            \begin{tabular}{p{2.6cm}p{1cm}p{1cm}p{1cm}}
            \toprule
            \textbf{Context-Query}  & \textit{Single} & \textit{Step=10}  & \textit{Step=100} \\
            \midrule
            Direct prompt &100.0 &99.0 &90.0 \\
            Equivalent prompt &74.0 &75.0 &64.0 \\
            \hdashline
            Cloze &77.7 &80.5 &75.5\\
            Related context &84.4 &81.0 &85.0 \\
            \quad reference &44.0 &37.0 & 48.0\\
            Raising doubts &20.7 &22.0 &21.5\\
		\bottomrule
	\end{tabular}
 \vspace{-1.5mm}
        \caption{Results on multiple-instance edit of MEMIT.}
	\label{multiedit}
 \vspace{-4.5mm}
\end{table}

The main observations are consistent with the current main conclusion:
Multiple-instance edit is also prone to our ``attacking'' prompts.
MEMIT performs well for multiple edits as claimed, while more edits still cause a lower overall performance. Expression changes hurt multiple edits of MEMIT more than related contexts.

\textbf{Baseline coverage.} 
From a principled perspective, robustness is a property of the editing method, not of the baseline LLM. 
To focus on communicative AI, the mainstream architecture of the most powerful open-source communicative AI is the decoder-only Transformer.
Some important editing methods are mainly for decoder-only Transformers (ROME) or large models (IKE), which makes the Llama family suitable. 
In our auxiliary experiments, observations on GPT-J-6B, Vicuna-7B, and ChatGLM-6B are consistent with our findings, i.e., the vulnerability to neighbor knowledge and complex forms.

\section{Temporal-based Knowledge}
\label{mquake}
Our motivation is expanded to a time-related benchmark for the scalability of our findings and enhancement of the motivation for practical editing. We consider MQAUKE-T \cite{zhong2023mquake}, the available knowledge edit benchmark to simulate the temporal knowledge update in the real world. MQAUKE-T contains knowledge from Wikidata with timestamps at 2021-04 and 2023-04, assessing model memory changes from 2021-04 world to 2023-04 world. GPT-J-6B, an LLM trained before 2023 is adopted to edit.
Representative attacking prompts are evaluated on ROME, MEMIT, and IKE. 

Table \ref{time} presents the results.
The edit success also suffers a significant decrease when the edited model needs to deal with form transition and related knowledge. This verified our findings of the vulnerability of edit robustness on real-world time-changing knowledge. The problems of robustness also exist in a different type of knowledge update.
\begin{table}[htb]
	\centering\small
            \begin{tabular}{p{2.3cm}|p{0.4cm}p{0.4cm}|p{0.4cm}p{0.4cm}|p{0.4cm}p{0.4cm}}
            \toprule
            \textbf{Editing Method}  & \multicolumn{2}{c|}{\textbf{ROME}} & \multicolumn{2}{c|}{\textbf{MEMIT}} & \multicolumn{2}{c}{\textbf{IKE}} \\
            \midrule
            \textbf{Context-Query}  & \textit{acc} & \textit{rev}  & \textit{acc} & \textit{rev}  & \textit{acc} & \textit{rev} \\
            \midrule
            Direct prompt &100.0 &-- &100 &-- &94.8 &-- \\
            Equivalent prompt &73.9 &9.4 &73.9 &6.3 &85.4 &0.0 \\
            \hdashline
            Cloze &37.0 &4.9 &25.3 &5.1 &55.7 &2.6 \\
            Related context &84.4 &6.3 &80.2 &10.4 &96.9 &2.1\\
            Raising doubts &46.3 &32.3 &42.7 &34.8 &2.1 &26.6\\
		\bottomrule
	\end{tabular}
 \vspace{-1.5mm}
        \caption{Results on MQAUKE-T of GPT-J-6B. \textit{acc}: \textit{accuracy}, \textit{rev}: \textit{reversion}. The \textit{Related context} means adding context to the direct prompt. Other denotations are consistent with Table \ref{main}.}
	\label{time}
 \vspace{-4.5mm}
\end{table}

\begin{figure*}[thb]
  \centering
    \setlength{\abovecaptionskip}{0cm}
    \setlength{\belowcaptionskip}{-0.3cm} 
  \subfigure[Perplexity distributions by Llama\quad \quad -2-7B-chat.]{
    \label{probeone} 
    \includegraphics[width=0.31\textwidth]{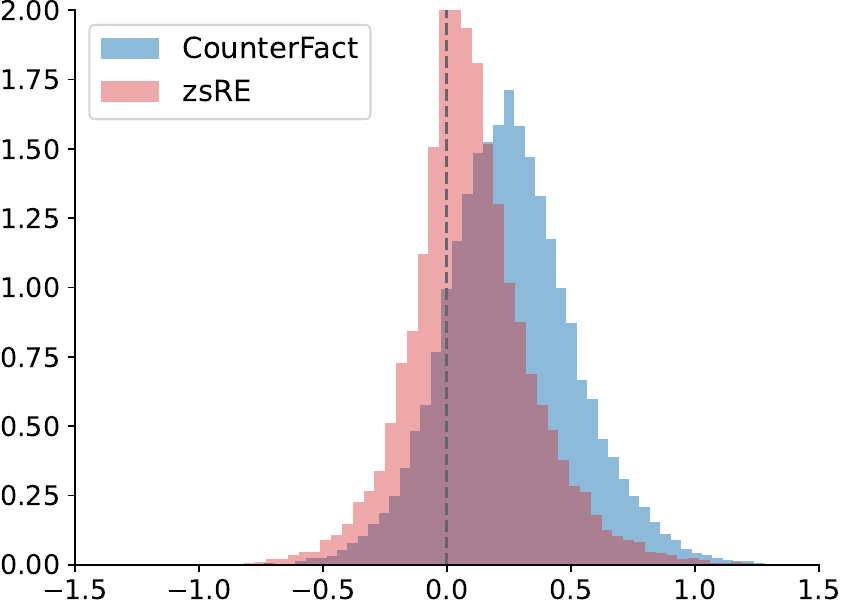}}
  \subfigure[Spearman correlation scores between the ICL accuracy and Frequency or Co-occurrence across relations types.]{
    \label{probetwo} 
    \includegraphics[width=0.63\textwidth]{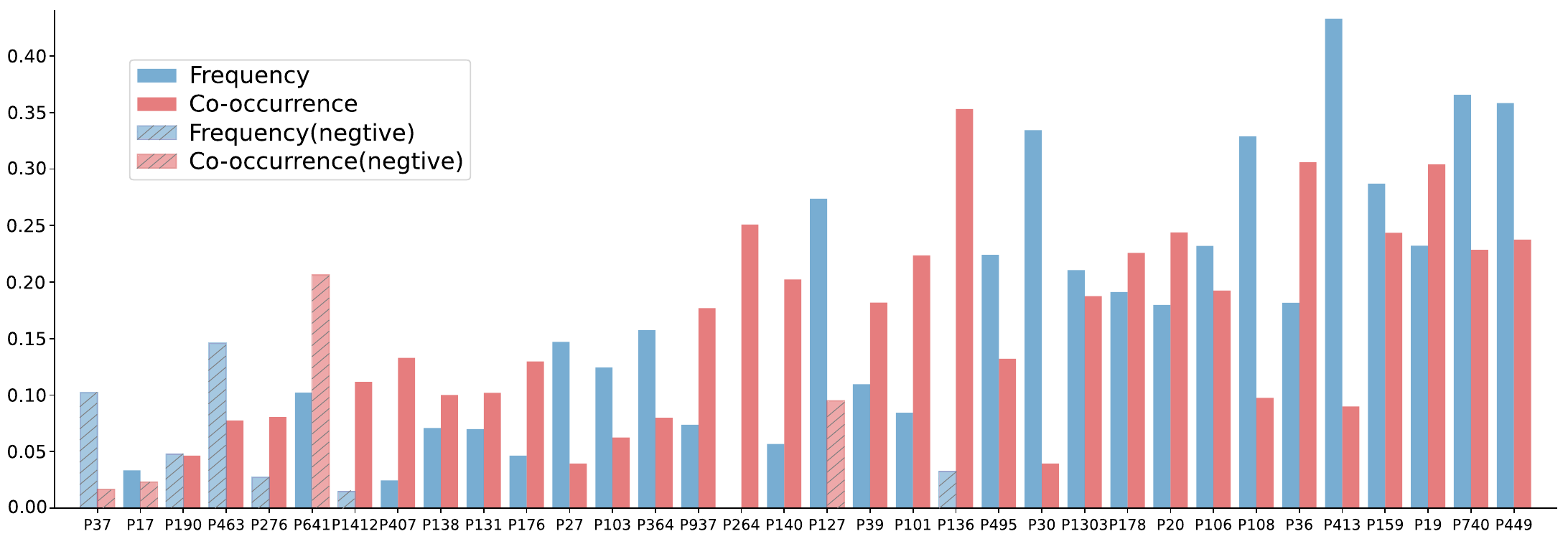}}
  \caption{Probe the knowledge in Llama through (a) perplexity and (b) prompt results.}
  \label{probe} 
\end{figure*}

\section{Details of Experiments for \textit{RQ3}}
\label{appc}
\subsection{Measurements Implementation}
The queries for the three measurements of knowledge features are as follows.

\noindent (i) Frequency. Following \citet{mallen-etal-2023-trust}, The URL is requested as 

\textit{\nolinkurl{https://wikimedia.org/api/rest\_v1/metrics/pageviews/per-article/en.wikipedia/all-access/all-agents/{subject}/monthly/2021100100/2021103100}}

\noindent (ii) Connection. The query to WikiData is 
\begin{verbatim}
SELECT (COUNT(?neighbor) AS ?edgeCount)
WHERE {
wd:{subject} ?p ?neighbor.
}
\end{verbatim}

\noindent (iii) Co-occurrence. The query to WikiData is 
\begin{verbatim}
SELECT (COUNT(*) AS ?pathCount)
WHERE {
{
    wd:{subject} ?p1 ?middle.
    ?middle ?p2 wd:{object}.
    FILTER (?middle != wd:{subject} &&
    ?middle != wd:{object})
}
}
\end{verbatim}

\subsection{Supplementary Figure}
Figure \ref{probe} (a) presents the distribution of the logarithmic perplexities difference of $o$ and $o^\prime$. There are 15.08\% samples in CounterFact and 35.65\% in zsRE whose original objects have no smaller perplexities than the new object.

Figure \ref{probe} (b) shows the correlation between knowledge popularity and parametric memory with Spearman correlation scores between ICL accuracy and Frequency or Co-occurrence on CounterFact. Most relation types have scores around 0.1$-$0.3. A few relation types are negative outliers.
For example, the relation \textit{[X] and [Y] are twin cities} rarely exists in memories and gets various outputs. The samples of relation \textit{[X] is a member of [Y]} always end with the same answer \textit{FIFA}.

\section{Experiments for Potential Mitigation}
\label{appmiti}
\subsection{Experiments and Results}
\begin{table}[htb]
	\centering\scriptsize
        \tabcolsep=0.15cm
            \begin{tabular}{p{1.5cm} p{0.3cm} p{0.3cm}p{0.4cm} p{0.3cm}p{0.4cm} p{0.3cm}p{0.4cm} p{0.3cm}p{0.4cm} }
            \toprule
            \textbf{Method}  &\textbf{BL} & \multicolumn{2}{c}{\textbf{+Samp.}} & \multicolumn{2}{c}{\textbf{+Disen.}} & \multicolumn{2}{c}{\textbf{+Disen.$^\dagger$}} & \multicolumn{2}{c}{\textbf{+Reso.}}\\
            \midrule
            \textbf{}  & \textit{acc}   & \textit{acc} & \textit{diff}  & \textit{acc} & \textit{diff}& \textit{acc} & \textit{diff}& \textit{acc}  & \textit{diff} \\
            \midrule
            Direct prompt &99.9  &100 &+0.1 &100 &+0.1 &100 &+0.1  &100 &+0.1 \\
            Cloze &67.0  &70.6  &+3.6 & 41.7 &$-$25.3 &70.4 &+3.4  &66.7 &$-$0.3  \\
            Related context &55.6  &71.8 &+16.2 &62.1 &+6.6 &74.0 &+18.4 &62.6 & +7.0\\
            \quad w/ reference &21.0 &29.0 &+8.0 &45.2 &+23.2 &67.0 &+46.0  &36.8 &+15.8 \\
            Raising doubts &16.9 &13.5 &$-$3.4 &75.5 &+58.6 &51.3 &+34.4 &16.9 &+0.0 \\
            \midrule
            Average &52.1 &57.0 &+4.9 &64.9 &+12.8 &72.5 &+20.5 &56.6 &+4.5 \\
		\bottomrule
	\end{tabular}
        \caption{Mitigation validation on ROME. BL means the baseline of the original ROME method.}
	\label{mitigation}
\end{table}

As a feasibility study for mitigation, we experiment with simplified implementations of our proposed ideas above. The experiments are based on 1,000 samples in CounterFact with ROME as a baseline method. We leave further improvement of robustness for future work.

Table \ref{mitigation} presents our results, where each method shows performance improvements on average.
For method (i), we add related contexts at the sampling step when computing the average target key-value pairs in ROME, dubbed as \textit{Samp.}. This mainly improves the scores on various contexts (i.e., Cloze, related context).
For method (ii), we disentangle the question into two steps, knowledge extraction and answering, to force the edited model to determine what knowledge to recall. This is dubbed as \textit{Disen.}. The disentanglement step helps ROME with long contexts and doubtful questions, while it also causes decreases in cloze. As local edits can hurt general abilities like reasoning \cite{gu2024model}, we try to call an LLM API (GLM-4 \cite{du2021glm}) for the knowledge extraction step, dubbed as \textit{Disen.$^\dagger$}, which leads to consistent increases.
For method (iii), as an example of targeted mitigations, we ask the edited model to rewrite the question if the subject is referred to by a pronoun, dubbed as reference resolution, \textit{Reso.}. This improves the scores for questions with reference.
Further studies on advanced editing methods are left for future work.
\subsection{Details}
\textbf{Method (i): sampling.}
Editing methods adopt a context sampling step for generalization. In the implementation of ROME, the parameter update requires the targeted hidden states before ($k*$) and after ($v*$) the edited MLP. At this step, the subject embedding is an average across prefix sampling. The prefixes are 20 texts, ten of length 5 and ten
of length 10, gathered by generating begin with very frequent words (``The'', ``Therefore'', ``Because'', ``I'', ``You''). We add related contexts, irrelevant contexts, and dialogues to those samples and truncate them to 100-token lengths.

\textbf{Method (ii): disentanglement.} The disentanglement is implemented by two-step prompting, adding a knowledge extraction step. The prompt template is shown below. Then the prompt and the output extraction are the input for the answer.
\begin{prompt}[title={Knowledge extraction prompt template.}]
Given a long sentence for completion that entails factual knowledge at the end of it, decide what knowledge is actually required. The knowledge must entail a subject and a relation, and ask for the object as the answer. 

Sentence: "Fill the blank. Q: Tony Iommi is well-known for performing \_\_. A:Guitar.
Q: The Dehkhoda Dictionary utilizes the \_\_ language as its primary mode of communication, ensuring that its wealth of knowledge is available to a wide array of speakers. A: "

Knowledge: "The language of Dehkhoda Dictionary is"

Sentence: "Human: Hey, what's up? I heard you know a lot about the Dehkhoda Dictionary. Can you tell me more about it?
Assistant: Sure! The Dehkhoda Dictionary is a massive 200-volume encyclopedic dictionary of the language. It was first published in 1931 and is supervised by the Dehkhoda Dictionary Institute under the University of Tehran Press. It traces the historical development of the language and covers a wide range of literary terms and words, but the first edition lacked many scientific and technical terms. Newer editions have been published yearly and include those terms. The language of Dehkhoda Dictionary is"

Knowledge: "The language of Dehkhoda Dictionary is"

Sentence: "The language of Dehkhoda Dictionary is Russian.??? No, It can't be, This contradicts my understanding. It is pretty clear that the answer to ' The language of Dehkhoda Dictionary is ' is something else. Rethink and give the correct answer."
Knowledge: "The language of Dehkhoda Dictionary is"

Sentence: "\{prompt\}"

Knowledge: 

\end{prompt}
\textbf{Method (iii): reference resolution.} Similar to the disentanglement, two-step prompting is triggered if the last sentence (question) in the input contains a pronoun that replaces the subject entity. We use the same prompt template as disentanglement but only used for a pronoun subject. 

\end{document}